\documentclass{article}

\PassOptionsToPackage{numbers}{natbib}

\usepackage[preprint]{neurips_2026}

\usepackage[utf8]{inputenc} % allow utf-8 input
\usepackage[T1]{fontenc}    % use 8-bit T1 fonts
\usepackage{hyperref}       % hyperlinks
\usepackage{url}            % simple URL typesetting
\usepackage{booktabs}       % professional-quality tables
\usepackage{amsfonts}       % blackboard math symbols
\usepackage{nicefrac}       % compact symbols for 1/2, etc.
\usepackage{microtype}      % microtypography
\usepackage{xcolor}         % colors
\usepackage{graphicx}
\usepackage{algorithm}
\usepackage{algpseudocode}
\usepackage{amsmath}
\usepackage{wrapfig}
\usepackage{booktabs} % for professional tables
\usepackage{tcolorbox}
\usepackage{tabularx}
\usepackage{float}
\usepackage{xurl}
\usepackage{hyperref}
\title{Training Reasoning Models on Saturated Problems
\\ via Failure-Prefix Conditioning}

\author{
Minwu Kim\thanks{Corresponding author: \texttt{mwk300@nyu.edu}} \quad
Safal Shrestha \quad
Anubhav Shrestha \quad
Keith Ross \\
New York University Abu Dhabi \\
}

\begin{document}

\maketitle

\setcounter{footnote}{0}

\begin{abstract}

As Reinforcement Learning with Verifiable Rewards (RLVR) substantially improves the reasoning abilities of large language models (LLMs), a new bottleneck emerges: more training problems become saturated, that is, the LLM answers the questions correctly for nearly every rollout. 
On such problems, rewards provide little useful learning signal. 
While collecting harder problems is a natural response, it is costly and increasingly difficult. 
We propose \textit{failure-prefix conditioning}, a simple method that unlocks the remaining signal in saturated problems by shifting exploration toward failure-prone reasoning states. By conditioning on prefixes of rare incorrect trajectories, 
the method improves the model’s ability to recover from misleading early reasoning.
We observe that failure-prefix conditioning consistently improves performance where standard RLVR stalls, and achieves gains comparable to training on newly collected medium-difficulty problems.
We further analyze the model's robustness, finding that our method reduces performance degradation under misleading failure prefixes, albeit with a mild trade-off in adherence to correct early reasoning.
% Additionally, performance is robust to the choice of target accuracy, and a lightweight fixed-truncation variant achieves similar results.
Finally, we demonstrate that an iterative approach, which refreshes failure prefixes during training, unlocks additional gains after performance plateaus. Overall, our results show that saturated problems still contain valuable learning signal, and that failure-prefix conditioning provides an effective way to unlock it. \footnote{Code available at \url{https://github.com/minwukim/training-on-saturated-problems}}

\end{abstract}

\section{Introduction}
Reinforcement Learning with Verifiable Rewards (RLVR) has substantially improved the reasoning capabilities of large language models (LLMs) \cite{guo2025deepseek, lambert2025tulu3, openai2024o1}. Today’s frontier models achieve near-ceiling scores on advanced benchmarks such as MATH and AIME \cite{akhtar2026benchmark_saturation} and attain gold-medal-level performance on the IMO, IOI and ICPC \cite{ openai2025competitive_programming, deepmind2025gemini_icpc, luong2025robust_math_reasoning}. Beyond such standardized evaluations, they are now producing results beyond established human knowledge, such as discovering a new matrix multiplication procedure \cite{novikov2025alphaevolve} and solving multiple previously unsolved Erd\H{o}s problems \cite{alexeev2026shortproofs2,feng2026gemini_erdos}.

% However, this progress also creates a new bottleneck. As models improve, more training problems become saturated. On such problems, rewards become almost deterministic and provide little useful learning signal as the model solves them correctly in nearly every rollout~\cite{yu2025dapo, zeng2025rlve}. Continuing progress would then seem to require a supply of harder, unsaturated problems, but manually collecting such problems only becomes more difficult and more expensive over time \cite{villalobos2024run_out_of_data}.

However, this progress also creates a new bottleneck. As models improve, more training problems become saturated, that is, the LLM answers the questions correctly for nearly every rollout. On such problems, rewards provide little useful learning signal~\cite{Kim2025RLvsDistill, yu2025dapo, zeng2025rlve}. Continuing progress would then seem to require a supply of harder, unsaturated problems, but manually collecting such problems only becomes more difficult and more expensive over time \cite{villalobos2024run_out_of_data}.

\begin{figure}[t]
    \centering
    \includegraphics[width=\linewidth]{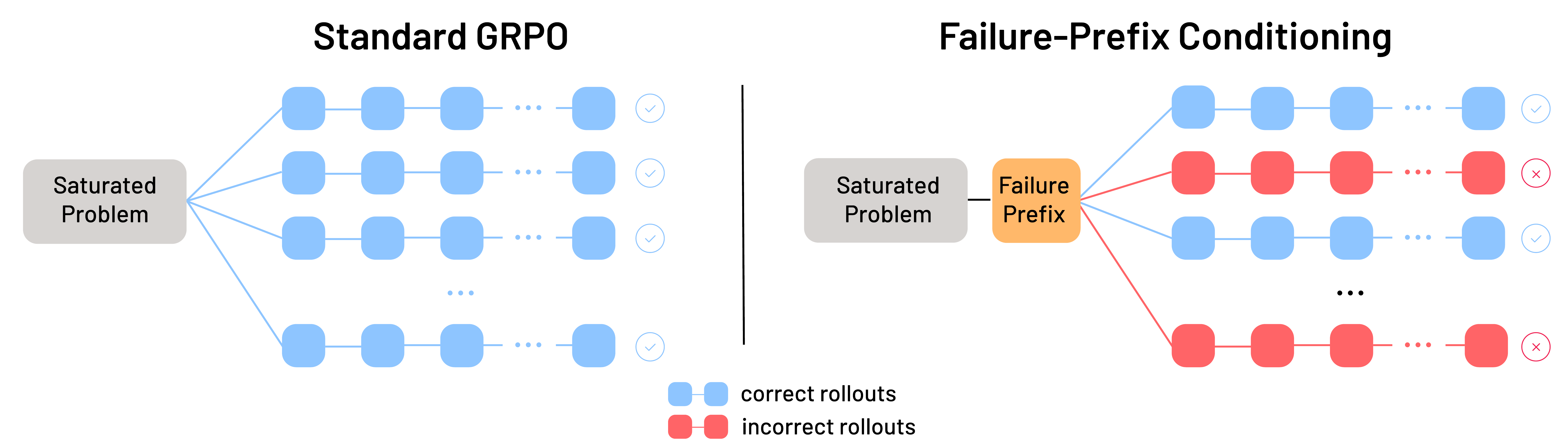}
    \vspace{-1.5em}
    \caption{Illustration of standard GRPO training and training with failure-prefix conditioning (ours). While standard GRPO predominantly generates correct rollouts, failure-prefix conditioning exposes the model to failure-prone reasoning states, making informative failures more accessible.}
    \label{fig:method}
    \vspace{-1.5em}
\end{figure}

Instead of collecting ever more challenging training data, we address \textit{how to extract as much learning signal as possible from the saturated training problems in the existing dataset}. Importantly, saturation does not imply that the learning signal is exhausted. Even on problems with near-perfect rollout accuracy, incorrect reasoning trajectories still exist, albeit extremely rarely under standard sampling. As established in prior work, such informative failures play a crucial role in training~\cite{setlur2025e3, zhu2025negative}. However, most of the sampling budget is spent generating redundant correct solutions, with failures becoming exceedingly sparse~\cite{wang2025oneexample}. This suggests that the core challenge in learning from saturated problems is not the \textit{absence of informative failures}, but their \textit{poor accessibility}. If failures could be encountered more frequently, RLVR could potentially continue to make progress even on saturated problems. 

To this end, we propose \textit{failure-prefix conditioning}, a simple and effective method for learning from saturated problems in RLVR. As shown in Figure~\ref{fig:method}, instead of sampling responses from the original question, this method explicitly targets failure-prone reasoning states.
We identify rare incorrect rollouts from saturated questions, and create new prompts consisting of the original question followed by a prefix of the incorrect rollout. 
We then select prefix lengths such that the resulting rollout accuracy is close to a target value $\tau$, which controls the difficulty of the resulting training examples. In practice, we set $\tau=0.5$, where the learning signal is maximized~\cite{li2025disco}.
% We select the specific prefix lengths that maximize the learning signal (i.e., target accuracy $\tau=0.5$)~\cite{li2025disco}. 
Training on the resulting dataset directs exploration toward failure-prone regions of the response space 
% , enabling effective learning from saturated problems.
and improves its ability to recover from misleading early reasoning.

We evaluate our method in a high-accuracy training regime, where many training questions are already saturated. This setting reflects the setting faced by today’s frontier reasoning models. Specifically, we use DeepSeek-R1-Distill-Qwen-1.5B~\cite{guo2025deepseek} as the base model and the MATH train set~\cite{hendrycks2021measuring} as the training data. We identify saturated questions, construct the failure-prefix-conditioned dataset, and train the model with RLVR. We compare against several baselines designed to isolate the effect of our method: standard RLVR on the full MATH training set, standard RLVR on the same saturated subset without conditioning, and standard RLVR on an equal number of medium-difficulty questions. These comparisons test whether failure-prefix conditioning extracts more learning signals from saturated problems and whether it can match the benefit of collecting new high-signal problems.

Across math reasoning benchmarks of varying difficulty, training on the full MATH set or directly on the saturated subset provides negligible improvement, confirming that standard RLVR stalls in this regime. In contrast, failure-prefix conditioning yields consistent gains over the base model, demonstrating that our method can effectively recover learning signal from saturated problems. We further show this improved performance is on par with that of instead training on an equal number of additional medium-difficulty questions. 
We also find that failure-prefix conditioning is robust to the exact target accuracy. Although $\tau=0.5$ performs best, nearby choices still yield strong gains. Motivated by this robustness, we consider a lower-cost \textit{fixed-truncation} variant: instead of estimating rollout accuracy for all candidate prefixes and selecting the one closest to $\tau$, it simply truncates each incorrect trajectory at a fixed fraction $\gamma$ of its length. This achieves comparable performance while substantially reducing the overhead in dataset construction.

Next, we evaluate how well our method improves the model's ability to recover from misleading reasoning. As compared with baseline models, models trained with failure-prefix-conditioned datasets are substantially better at recovering from misleading reasoning. When given incorrect intermediate steps, their performances degrade more slowly, showing stronger robustness to early mistakes. We also observe a mild trade-off: when given correct intermediate steps, these models show a smaller accuracy increase than the baselines, showing tendency to diverge more from correct early reasoning. Overall, however, the robustness gains dominate, leading to improved aggregate performance.

Finally, we explore whether learning from saturated problems can be further extended by iteratively refreshing failure prefixes as the model improves. As training progresses, performance under a fixed failure-prefix-conditioned dataset eventually plateaus, suggesting that previously informative prefixes may become less effective as the policy updates. Therefore, we perform a second iteration of failure-prefix conditioning, resampling new failures from the updated model and reconstructing the prefix-conditioned dataset. Empirically, this iterative procedure yields 
% modest
additional gains beyond the initial plateau, suggesting that periodically updating failure prefixes may further recover additional learning signal and enable continued improvement.

We summarize our contributions as follows:
\begin{itemize}
    \item We propose failure-prefix conditioning, a method that directs exploration toward failure-prone reasoning states to enable effective RLVR training on saturated problems.
    \item We show that failure-prefix conditioning enhances robustness to incorrect intermediate reasoning trajectories, enabling better recovery from misleading partial solutions compared to standard RLVR baselines.
    \item We demonstrate that iteratively refreshing failure prefixes can yield additional gains after performance plateaus, offering a potential pathway for further extending learning on saturated problems.
\end{itemize}
Overall, our results establish that saturated problems still contain valuable learning signal, and that failure-prefix conditioning is an effective method for unlocking their potential.

\section{Preliminaries: RLVR and Saturated Problems}\label{sec:preliminary}
In this section, we explain why training with GRPO~\cite{shao2024deepseekmath}, a standard RLVR algorithm, stalls on saturated problems. Section \ref{subsection2.1} reviews the training procedure and notation, and Section \ref{subsection2.2} analyzes why saturation leads to stalled learning.

\subsection{GRPO Training} \label{subsection2.1}

 Let $\mathcal{D} = \{(q, a^\ast)\}$ denote a dataset of questions $q$ with verifiable ground-truth answers $a^\ast$.
Given a policy $\pi_\theta$, we generate $N$ independent rollouts
\[
y_i \sim \pi_\theta(\cdot \mid q), \quad i = 1, \ldots, N,
\]
and assign each rollout a binary reward
$r_i = r(y_i; q)$,
where $r_i = 1$ if $y_i$ provides the correct answer $a^\ast$ and $r_i = 0$ otherwise. For each rollout in $\{y_i\}_{i=1}^N$, the advantage is given by
$A_i = (r_i - \mu_r)/(\sigma_r + \epsilon)$,
where $\mu_r = \frac{1}{N}\sum_{i=1}^N r_i$, $\sigma_r = \mathrm{std}(\{r_i\}_{i=1}^N)$, and $\epsilon > 0$ is a small constant that prevents division by zero when $\sigma_r = 0$.

The policy $\pi_\theta$ is then updated by minimizing the following loss, where we omit the clipping and KL-regularization terms for clarity:
\[
\mathcal{L}_{\text{GRPO}}(\theta)
= - \mathbb{E}_{(q,a^\ast)\sim\mathcal{D}}
\left[
\frac{1}{N}\sum_{i=1}^N A_i \log \pi_\theta(y_i \mid q)
\right].
\]

\subsection{Training with Saturated Problems}\label{subsection2.2}
We say that a problem $q$ is \emph{saturated} under policy $\pi_\theta$ if the rollout accuracy $p_\theta(q)$ is close to 1, where
\[
p_\theta(q) := \Pr_{y \sim \pi_\theta(\cdot \mid q)}[r(y;q)=1]
\]
For such a problem, with high probability all sampled rollouts satisfy $r_i = 1$, yielding
\[
\mu_r = 1, \quad \sigma_r = 0,
\]
and consequently $A_i = 0$ for all $i$.
As a result, the policy gradient vanishes and training stalls.

Even when a rare incorrect rollout occurs, reward variance remains small.
For binary rewards,
\[
\mathrm{std}[r] = \sqrt{p_\theta(q)(1 - p_\theta(q))},
\]
which approaches zero as $p_\theta(q) \to 1$. Since the policy gradient magnitude scales with this standard deviation value~\citep{li2025disco} (see Appendix~\ref{app:weight-variance-derivation} for a full derivation), learning signals on saturated problems are weak. This analysis is also empirically supported by the findings of \citet{zhan2025exgrpo}, who observe substantially smaller gains from training on easy questions ($p_\theta(q)\in[0.75,1.00)$) compared to moderately difficult ones ($p_\theta(q)\in(0.25,0.75]$).

\section{Methodology}
We propose \textit{failure-prefix conditioning}, a simple and effective method for learning from saturated problems in RLVR. The core idea is to reshape exploration by conditioning training on failure-prone reasoning states, instead of initiating exploration from the original question. Section~\ref{subsection3.1} provides the motivation for our approach, and Section~\ref{subsection3.2} presents the proposed methodology.

\subsection{Motivation}
\label{subsection3.1}
As analyzed in Section \ref{subsection2.2}, 
learning stalls on saturated problems not because the model is perfect, but because its failures are too rare under standard RLVR.
Even when rollout accuracy is close to $1$, incorrect responses still exist in the model’s response space. However, during training, these failure-prone states are encountered too rarely to provide meaningful learning signal. In other words, the fundamental limitation is not the \textit{absence of informative failures}, but their \textit{poor accessibility}. 

To address this, rather than repeatedly sampling from the question as in the standard setting, we target exploration toward regions of the response space where uncertainty is higher and failures occur more frequently. Concretely, we propose \textit{failure-prefix conditioning}: condition training on prefixes obtained from rare incorrect responses, enabling RLVR to begin exploration directly from failure-prone states and improving the model’s ability to recover from misleading early reasoning.

This simple change dramatically improves accessibility. Consider a saturated problem with rollout accuracy $p_\theta(q)\approx0.95$. Under standard sampling, failures occur with probability $0.05$, meaning only a small number of incorrect responses are expected, and most rollouts are redundant successes that contribute little learning signal. If, instead, training begins from a failure-prone state where, say, $p\approx0.50$, the same sampling budget yields substantially more incorrect responses. As a result, informative failures become far more accessible.

\subsection{Methodology: Failure-Prefix Conditioning}
\label{subsection3.2}
% \begin{algorithm}[t] 
% \caption{Failure-Prefix Conditioning}
% \label{alg:failure-prefix-conditioning}
% \begin{algorithmic}[1]
% \REQUIRE Saturated dataset $\mathcal{D}_{\text{saturated}} = \{(q, a^\ast)\}$, policy $\pi_\theta$, target accuracy $\tau = 0.5$, number of rollouts $N$
% \ENSURE Prefix-conditioned dataset $\mathcal{D}'$

% \STATE Initialize $\mathcal{D}' \leftarrow \emptyset$
% \FOR{each $(q, a^\ast) \in \mathcal{D}_{\text{saturated}}$}
%     \STATE Obtain an incorrect rollout $\tilde{y} \sim \pi_\theta(\cdot \mid q)$
%     \STATE Let $L \leftarrow |\tilde{y}|$
%     \STATE Construct prefix set
%     \[
%         \mathcal{S}(q) = \{\tilde{y}_{1:\alpha_k}\}_{k=1}^K,
%     \]
%     \FOR{each prefix $s \in \mathcal{S}(q)$}
%         \STATE Sample $N$ rollouts $\{y_i\}_{i=1}^N \sim \pi_\theta(\cdot \mid q \oplus s)$
%         \STATE Estimate rollout accuracy
%         \[
%         p_\theta(q,s) \leftarrow \frac{1}{N}\sum_{i=1}^N \mathbb{I}[r(y_i;q)=1]
%         \]
%     \ENDFOR
%     \STATE Select prefix
%     \[
%     s^{(q)} \leftarrow \arg\min_{s \in \mathcal{S}(q)} \left|p_\theta(q,s) - \tau \right|
%     \]
%     \STATE Add $(q \oplus s^{(q)}, a^\ast)$ to $\mathcal{D}'$
% \ENDFOR
% % \RETURN $\mathcal{D}'$
% \end{algorithmic}
% \end{algorithm}

Let  $\pi_\theta$ be a fixed policy and $\mathcal{D}_{\text{saturated}}$ be a set of saturated questions, defined as questions whose rollout accuracy $p_\theta(q)$ is close to $1$.
%under standard sampling, as in Section~\ref{subsection2.2}. 
For each such question $q$, we obtain at least one incorrect rollout $\tilde{y}$ produced by $\pi_\theta$.
From each incorrect rollout $\tilde{y}$, we then construct a set of prefixes
\[
\mathcal{S}(q) = \{\tilde{y}_{1:\alpha_k}\}_{k=1}^K,
\]
where $\tilde{y}_{1:\alpha_k}$ denotes the first $\alpha_k$ tokens of $\tilde{y}$. In our experiments in Section~\ref{sec:experiment}, we sweep through prefix lengths corresponding to $10\%, 20\%, \ldots, 90\%$ of the total trajectory length, yielding $K=9$ candidate prefixes per question.

Each prefix $s \in \mathcal{S}(q)$ defines a prefix-conditioned prompt $q \oplus s$, from which we estimate the prefix-conditioned rollout accuracy
\[
p_\theta(q, s) = \Pr_{y \sim \pi_\theta(\cdot \mid q \oplus s)}[r(y;q)=1]
\]
using a fixed number of rollouts. Intuitively, when generation starts from the original question $q$, the rollout accuracy is near one for saturated problems. As progressively longer prefixes of an incorrect trajectory are prefixed, the conditioning context increasingly constrains the model toward failure states, and the rollout accuracy tends to decay towards $0$~\cite{wen2025parathinker}. While this decay is not necessarily monotonic, sweeping over prefix lengths reliably yields intermediate prefixes whose rollout accuracy lies between these extremes.

We select a single prefix $s^{(q)}$ whose rollout accuracy is closest to a target value $\tau \in (0,1)$:
\[
s^{(q)} = \arg\min_{s \in \mathcal{S}(q)} \left| p_\theta(q,s) - \tau \right|.
\]
In our main experiments, we set $\tau = 0.5$, where binary rewards exhibit the highest variance and GRPO provides the strongest learning signal~\cite{li2025disco}.
% and consider alternative values in ablation studies.

Using the selected prefixes, we construct a new training set
\[
\mathcal{D}' = \{(q \oplus s^{(q)}, a^{*}) \mid q \in \mathcal{D}_{\text{saturated}}\}.
\]
GRPO training is then performed on $\mathcal{D}'$ following the standard procedure described in Section~\ref{subsection2.1}.

% At first glance, this method may appear to introduce a train--test mismatch, since training rollouts start from prefix-conditioned prompts $q \oplus s^{(q)}$ whereas test-time generation starts from the original question $q$. We address this concern in Section~\ref{sec:train-test-mismatch}.

\section{Main Experiment} \label{sec:experiment}
\subsection{Experiment Settings}\label{subsec:experiment_setting}

% \begin{wrapfigure}{r}{0.35\textwidth}
%     \centering
%     \vspace{-3.5em}
%     \includegraphics[width=0.35\textwidth]{figures/MATH_distribution.pdf}
%     \vspace{-2em}
%     \caption{Distribution of rollout accuracy on the MATH train set.}
%     \label{fig:rollout-accuracy}
%     \vspace{-1em}
% \end{wrapfigure}

\textbf{Model and Dataset.} We consider a setting where the base model already achieves high accuracy on the training data, with many training questions are already saturated. This mirrors the regime faced by today's frontier reasoning models. Specifically, we use DeepSeek-R1-Distill-Qwen-1.5B~\cite{guo2025deepseek} as the base model and the MATH \cite{hendrycks2021measuring} train set of 7.5k questions as the training data. We estimate question-level rollout accuracy by generating 128 rollouts per question. The base model obtains 82.6\% accuracy on the training set, with the distribution shown in Appendix~\ref{appendix:train-set}. 

\textbf{Failure-prefix conditioning.} We divide the questions into 3 groups: (1) \emph{128/128}: fully saturated questions for which no incorrect rollout is observed; (2) \emph{121--127/128}: saturated questions with rollout accuracy greater than 120/128 and at least one observed incorrect rollout; and (3) \emph{0--120/128}: unsaturated questions with rollout accuracy of $\leq$120/128. These groups contain 2,455, 2,147, and 2,898 questions, respectively. We sample one incorrect rollout from each question in the 121--127/128 group and apply failure-prefix conditioning following Algorithm~\ref{alg:failure-prefix-conditioning} with target accuracy \(\tau = 0.5\), where reward variance is maximized. We also consider \(\tau=0.25\) and \(\tau=0.75\) for ablation studies. The cutoff of $120/128$ $(=15/16)$ is chosen to match our RLVR training configuration, where each question is sampled 16 times. Since an accuracy of $15/16$ is the largest value at which one expects at least one incorrect rollout in 16 independent samples, we treat questions above this level as saturated. Additional inference details are provided in Appendix~\ref{appendix:inference_details}.

\textbf{Baselines.} We train the base model on failure-prefix-conditioned datasets with different \(\tau\) values. We compare against the six baselines, each designed to examine a specific factor. 
\begin{enumerate}
    \item \emph{Base}: The base model.
    \item \emph{Full MATH}: Standard RLVR on the full MATH train set, which tests whether our method improves over conventional training on all available data. 
    \item \emph{Saturated}: Standard RLVR on the 121--127/128 questions, isolating the effect of failure-prefix conditioning versus standard training on the same data. 
    \item  \emph{Medium}: Standard RLVR on the same number of 2,147 medium-difficulty questions sampled from DeepScaleR~\cite{deepScaler2025} (a harder dataset than MATH), selected to have rollout accuracy 14--18/32 ($\approx50\%$) under the base model, where binary reward variance maximizes. 
    This evaluates whether failure-prefix conditioning can match or exceed the benefit of collecting an equal number of new high-learning-signal questions.
    \item \emph{Failure-Prefix + Unsaturated}: Failure-prefix-conditioned data (\(\tau = 0.5\)) combined with 0--120/128 questions.
    \item \emph{Unsaturated}: Standard RLVR on the 0--120/128 questions alone, isolating the contribution of unsaturated data.
    These last two baselines examine whether our method provides complementary gains when paired with unsaturated problems in the train set. 
\end{enumerate}

\textbf{RLVR training.} We train the base model using Dr.GRPO~\cite{liu2025r1zero}, a variant of GRPO that mitigates response length bias. All models use the same configuration with 16 rollouts per question, representing a standard, non-aggressive exploration setting. Further details are provided in Appendix~\ref{appendix:GRPO_training}.

\textbf{Evaluation.} We evaluate all models on five math reasoning benchmarks spanning a wide range of difficulties (from easiest to hardest): MATH500~\cite{hendrycks2021measuring}, AMC12~\cite{amc12dataset}, AIME24~\cite{aime24dataset}, AIME25~\cite{aime25dataset}, and HMMT25~\cite{hmmt25dataset}. For each question, we generate 32 samples and report the mean accuracy 
% (pass@1) 
to ensure statistical robustness. 
% Additionally, to determine if the method is merely sharpening the distribution~\cite{yue2025limit-of-rlvr, Kim2025RLvsDistill} or genuinely improving capability~\cite{du2025prorl, setlur2025e3}, we report pass@$k$ metrics up to pass@32 using the 32 responses. 
% All evaluations are performed with a maximum inference token limit of $32000$. 
Details of the evaluation setup are provided in Appendix~\ref{appendix:inference_details}.

\begin{table*}[t]
\centering
\caption{Performance on math benchmarks. Numbers in parentheses indicate absolute improvements over the base model. Best results are shown in \textbf{bold}, and second-best results are \underline{underlined}.}
\resizebox{\textwidth}{!}{
\begin{tabular}{l l l l l l l}
\toprule
\textbf{Model} & \textbf{MATH 500} & \textbf{AMC 12} & \textbf{AIME 24} & \textbf{AIME 25} & \textbf{HMMT 25} & \textbf{Average} \\
\midrule
Base
& 83.5 & 53.1 & 28.4 & 24.0 & 14.2 & 40.6 \\

Full MATH
& 84.5 (+1.0) & 54.6 (+1.5) & 27.8 (-0.6) & 24.2 (+0.2) & 13.6 (-0.6) & 40.9 (+0.3) \\

Saturated
& 84.3 (+0.8) & 52.7 (-0.4) & 29.0 (+0.6) & 24.0 (+0.0) & 13.5 (-0.7) & 40.7 (+0.1) \\

Medium
& 85.7 (+2.2) & 58.8 (+5.7) & 34.2 (+5.8) & 26.2 (+2.2) & 15.0 (+0.8) & 44.0 (+3.4) \\

Failure-Prefix + Unsaturated
& 85.7 (+2.2) & 55.6 (+2.5) & 30.7 (+2.3) & 25.8 (+1.8) & 13.8 (-0.4) & 42.3 (+1.7) \\

Unsaturated
& 86.0 (+2.5) & 56.5 (+3.4) & 26.0 (-2.4) & 25.7 (+1.7) & 14.2 (+0.0) & 41.7 (+1.1) \\

\midrule

Failure-Prefix ($\tau=0.5$)
& \textbf{86.6} (+3.1) & \textbf{60.5} (+7.4) & 33.1 (+4.7) & \underline{27.1} (+3.1) & \textbf{15.4} (+1.2) & \textbf{44.5} (+3.9) \\

Failure-Prefix ($\tau=0.25$)
& 85.6 (+2.1) & 58.2 (+5.1) & \textbf{35.5} (+7.1) & \textbf{27.7} (+3.7) & 14.2 (+0.0) & 44.2 (+3.6) \\

Failure-Prefix ($\tau=0.75$)
& 85.6 (+2.1) & 57.8 (+4.7) & 35.0 (+6.6) & 26.2 (+2.2) & 15.1 (+0.9) & 43.9 (+3.3) \\

\midrule

Fixed Truncation ($\gamma=0.25$)
& \underline{86.4} (+2.9) & \underline{59.3} (+6.2) & 32.8 (+4.4) & 26.4 (+2.4) & 14.6 (+0.4) & 43.9 (+3.3) \\

Fixed Truncation ($\gamma=0.5$)
& 86.4 (+2.9) & 58.5 (+5.4) & 34.3 (+5.9) & 26.9 (+2.9) & \underline{15.3} (+1.1) & \underline{44.3} (+3.7) \\

Fixed Truncation ($\gamma=0.75$)
& 85.7 (+2.2) & 58.3 (+5.2) & \underline{35.3} (+6.9) & 26.6 (+2.6) & 14.5 (+0.3) & 44.1 (+3.5) \\

\bottomrule
\end{tabular}
}
\label{tab:main_results}
\end{table*}

\subsection{Results}\label{sec:main_results}

\textbf{Failure-prefix conditioning enables learning from saturated problems.} Table~\ref{tab:main_results} shows that our method consistently improves performance across all evaluated math benchmarks from the easiest benchmark (MATH500) to the most challenging one (HMMT25). The average accuracy rises from 40.6\% for the base model to 44.5\%, corresponding to a +3.9 point gain. In contrast, standard RLVR on the full MATH training set (40.9\%) and on the same saturated questions without conditioning (40.7\%) yields little improvement, consistent with the saturation analysis in Section~\ref{subsection2.2}. These results show that failure-prefix conditioning recovers useful learning signal from saturated problems that standard RLVR cannot effectively exploit.

\textbf{Failure-prefix conditioning matches the benefit of new medium-difficulty data.} Medium-difficulty problems are effective for RLVR because its rollout accuracy is close to 50\%, where learning signal is maximized. Nevertheless, our method slightly outperforms training on medium-difficulty questions on average (44.5\% vs.\ 44.0\%), despite training only with the original saturated MATH questions. This indicates that failure-prefix conditioning can extract a learning signal from saturated questions that is comparable to, or even stronger than, training on newly collected unsaturated medium-difficulty problems. Additional experiments addressing response-length effects are provided in Appendix~\ref{sec:length}.

\textbf{Failure-prefix conditioning is robust to the target accuracy.} We set the main target accuracy to $\tau=0.5$ since this theoretically maximizes reward variance and thus the expected gradient signal, as discussed in Section~\ref{subsection3.2}. To test sensitivity to this hyperparameter, we repeat training with $\tau=0.25$ and $\tau=0.75$. As shown in Table~\ref{tab:main_results}, both variants remain effective: $\tau=0.25$ reaches 44.2\% (+3.6), and $\tau=0.75$ reaches 43.9\% (+3.3), only slightly below $\tau=0.5$ at 44.5\%. This implies that, while targeting maximal variance performs best, our method does not necessarily require precise tuning of the target accuracy. This result motivates a lower-cost variant of our method in Section~\ref{sec:fixed_truncation}.

\textbf{Combining unsaturated data with failure-prefix conditioning is not always beneficial.} In practical RLVR training pipelines, it is natural to use all available unsaturated problems. However, as shown in Table~\ref{tab:main_results}, combining failure-prefix-conditioned data with the 0--120/128 unsaturated subset reaches only 42.3\%, which improves over the base model by +1.7 points but remains well below failure-prefix conditioning alone (44.5\%). Meanwhile, training on the unsaturated subset alone reaches 41.7\% (+1.1), indicating that these examples are not deteriorating model performance by themselves. Rather, the results suggest that simple data mixing can dilute the benefit of failure-prefix conditioning. A possible reason is that many unsaturated examples may quickly become saturated during training and contribute weaker signals later on, though further study is needed. Given this, we conduct the remaining analyses using the failure-prefix-conditioned dataset alone.

\subsection{Reducing Prefix Selection Overhead via Fixed Truncation}
\label{sec:fixed_truncation}
A major source of computational overhead in our method comes from prefix length selection. Matching a target accuracy requires slicing each incorrect response at multiple lengths and estimating rollout accuracy for each candidate prefix, which is expensive (Appendix~\ref{appendix:overhead}). However, Section~\ref{sec:main_results} shows that performance is insensitive to the exact target value $\tau$, motivating a cheaper alternative.

We try \textit{fixed truncation}, where each incorrect response is truncated at a fixed fraction $\gamma$ of its length, without estimating prefix-conditioned accuracy. Specifically, we test $\gamma \in \{0.25, 0.5, 0.75\}$ following Algorithm~\ref{alg:fixed-truncation}. As shown in Table~\ref{tab:main_results}, we obtain average accuracies of 43.9\%, 44.3\%, and 44.1\%, respectively. These are only slightly below the 44.5 achieved with explicit $\tau=0.5$ matching. This small gap suggests that precise prefix-difficulty calibration is not necessary, and that fixed truncation offers a practical variant that significantly reduces overhead while preserving most of its benefits.

% \section{Why Failure-Prefix Conditioning Works}
\section{Recovery From Misleading Reasoning}\label{subsection:recovery}

\begin{figure}[t]
    \centering
    \includegraphics[width=\linewidth]{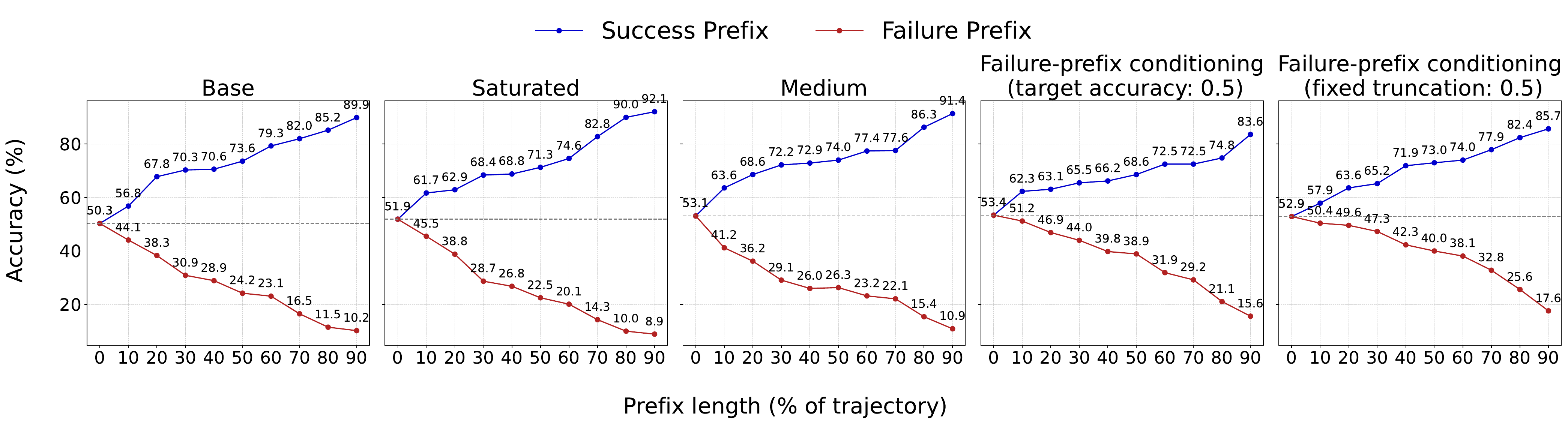}
    \vspace{-1.5em}
    \caption{Rollout accuracy versus prefix length (\% of trajectory) when conditioning on correct and incorrect prefixes.}
    \label{fig:recovery}
    % \vspace{-20pt} % move figure up a bit
\end{figure}

% \begin{figure} [h]
%     \centering
%     \includegraphics[width=0.6\linewidth]{figures/round2.pdf}
%     \vspace{-0.7em}
%     \caption{Effect of iterative failure-prefix conditioning on training dynamics. \emph{Left:} Training reward curves for prolonged iteration 1 (steps 0–1200) and iteration 2 that forks at step 600 and proceeds with updated failure prefixes through step 1200. \emph{Right:} Mean accuracy across benchmarks for both iterations measured from steps 600–1200, with peak performance point for each model highlighted.}
%     \label{fig:round2}
    
% \end{figure}

Failure-prefix conditioning directs exploration toward failure-prone regions of the response space, enabling the model to better recover from early mistakes. In this section, we evaluate how well our method improves recovery from misleading early reasoning.

Recall that, in Section~\ref{sec:main_results}, we evaluate each model by generating 32 responses for every MATH500 question. We reuse the same rollouts to analyze recovery behavior. We compare five models: \emph{Base}, \emph{Saturated}, \emph{Medium}, and two variants of our method: (i) \emph{FP-target}, which selects prefixes to match target accuracy $\tau=0.5$, and (ii) \emph{FP-fixed}, which uses a fixed truncation ratio $\gamma=0.5$.

We select the subset of questions for which all models produce at least one correct and one incorrect response, yielding 131 questions. This ensures that recovery is feasible for all models. For each model and question, we sample one incorrect response and construct prefixes at 10\%, 20\%, …, 90\% of its length. Conditioning on each prefix, we generate 32 continued rollouts and compute rollout accuracy (details in Appendix~\ref{appendix:inference_details}). We group and average rollout accuracies by the percentage values and analyze how accuracy degrades as the prefix extends further into an incorrect reasoning trajectory.

\subsection{Failure-prefix conditioning improves robustness to misleading prefixes.} 

As shown in Figure~\ref{fig:recovery}, all models exhibit declining accuracy as the prefix length increases, consistently showing the “tunnel vision” issue in LLM reasoning~\cite{wen2025parathinker}. However, FP-target and FP-fixed degrade substantially more slowly. At a 30\% prefix, FP-target drops by 6.5 points (53.4 to 46.9) and FP-fixed by 3.3 points (52.9 to 49.6), compared to 12.0 points for the base model (50.3 to 38.3), 13.1 points for the saturated model (51.9 to 38.8) and 16.9 points for the medium model. This gap persists across prefix lengths, indicating stronger robustness to misleading intermediate states. 

Notably, the medium model—despite having similar overall accuracy—shows significantly sharper degradation than the base and saturated models. This suggests that improved robustness is specific to failure-prefix conditioning, rather than a byproduct of higher overall performance.

\subsection{A mild trade-off on correct prefixes.} 

Additionally, we examine whether failure-prefix conditioning induces excessive deviation from correct reasoning when given an initially correct partial solution, which is an undesired behavior. To assess this, we also perform an identical experiment but with prefixes from a correct response.

Figure~\ref{fig:recovery} shows that the FP models benefit less from correct prefixes. For example, at 30\% prefix, FP-target improves by 12.1 points (53.4 to 65.5) and FP-fixed by 12.3 points (52.9 to 65.2), compared to 20.0 points (50.3 to 70.3) for the base, 16.5 points (51.9 to 68.4) for the saturated, and 19.1 points for the medium model (53.1 to 72.2). This pattern holds across the entire range of prefix lengths.

Overall, these results reveal a mild trade-off: failure-prefix conditioning substantially improves robustness to misleading reasoning, while slightly reducing gains from correct prefixes. Addressing this behavior may require further investigation in future work. Importantly, this effect is limited in magnitude and does not outweigh the robustness gains achieved under failure-prefix conditioning, as reflected in the overall performance improvements reported in Section~\ref{sec:main_results}.

\section{Iterative Failure-Prefix Conditioning}
\label{sec:round2}
% \begin{figure} [h]
%     \centering
%     \includegraphics[width=0.6\linewidth]{figures/round2.pdf}
%     \vspace{-0.7em}
%     \caption{Effect of iterative failure-prefix conditioning on training dynamics. \emph{Left:} Training reward curves for prolonged iteration 1 (steps 0–1200) and iteration 2 that forks at step 600 and proceeds with updated failure prefixes through step 1200. \emph{Right:} Mean accuracy across benchmarks for both iterations measured from steps 600–1200, with peak performance point for each model highlighted.}
%     \label{fig:round2}
    
% \end{figure}

% \vspace{-11pt} % move figure up a bit
% \vspace{-0.4em}
\begin{wrapfigure}{r}{0.57\textwidth}
    \centering
    \vspace{-1.9em}
    \includegraphics[width=0.57\textwidth]{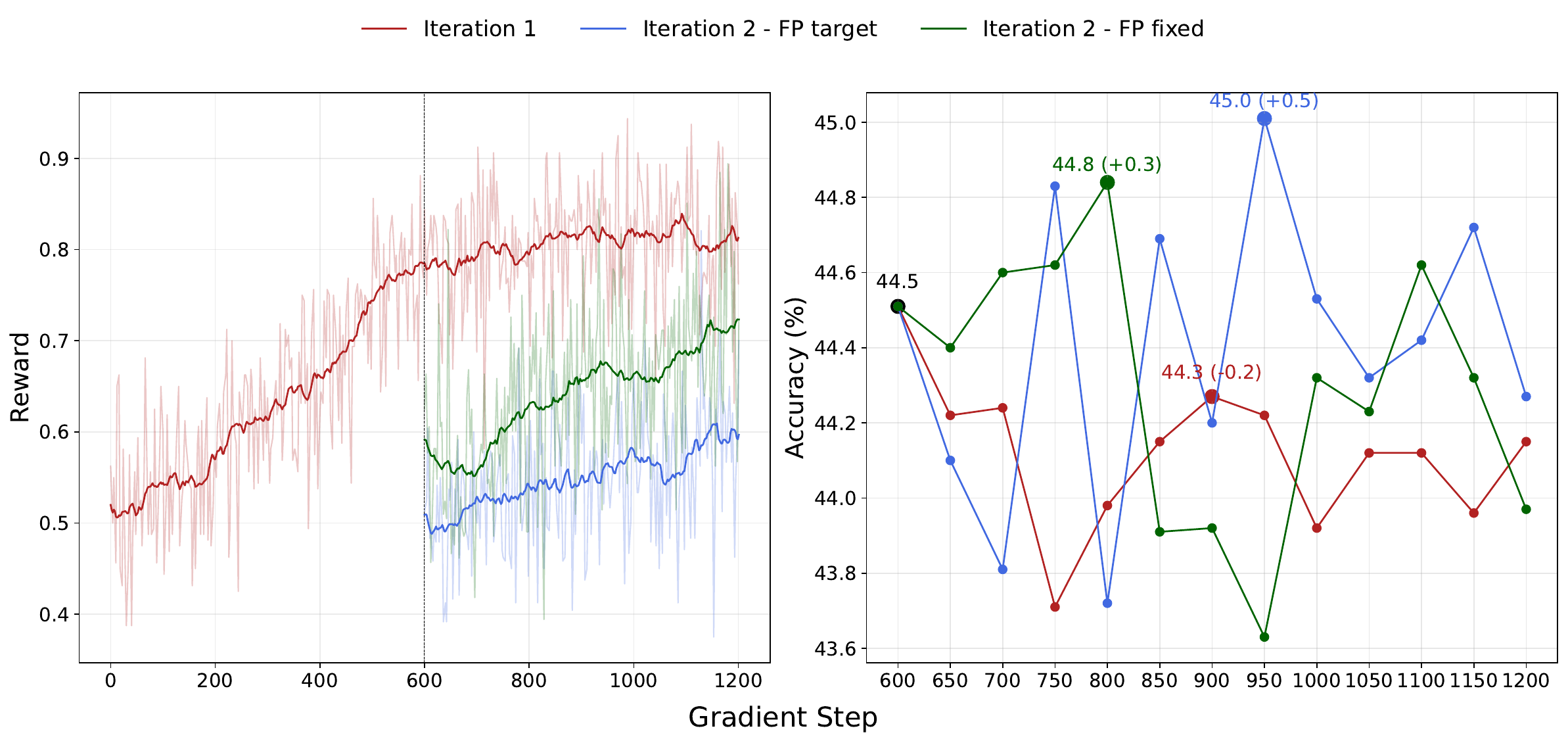}
    \vspace{-2em}
    % \caption{Training dynamics of iterative failure-prefix conditioning. \emph{Left:} Training reward curves. \emph{Right:} Mean accuracy across benchmarks 
    % % with peak performance point for each model highlighted
    % .}
    
    \caption{Training dynamics of iterative failure-prefix conditioning. \emph{Left:} Training reward curves for prolonged iteration 1 (steps 0–1200) and iteration 2 with updated failure prefixes through step 1200. \emph{Right:} Mean accuracy across benchmarks for both iterations measured from steps 600-1200, with peak performance point for each model highlighted.}    
    \label{fig:round2}
    \vspace{-1.0em}
\end{wrapfigure}

We observe that performance eventually plateaus as training with failure-prefix conditioning ($\tau=0.5$) progresses, with the mean reward flattening (Figure~\ref{fig:round2}, left). This behavior may arise from diminishing learning signal in the conditioned dataset, or from fixed failure prefixes becoming increasingly off-policy as the model improves.

To test whether additional gains can be obtained, we perform a second iteration of failure-prefix conditioning. Starting from the best checkpoint at step 600, we resample 128 responses for each of the MATH training questions. From these, we identify 2,124 questions with rollout accuracy 121--127/128. For each, we construct two failure-prefix-conditioned datasets: \emph{FP-target}, which selects prefixes to match a target accuracy $\tau=0.5$, and \emph{FP-fixed}, which uses a fixed truncation ratio $\gamma=0.5$. We then continue training from the checkpoint for an additional 600 gradient steps on each dataset.

We find that this second iteration yields further improvements. As shown in Figure~\ref{fig:round2} (right), extending training in the first iteration does not surpass the peak accuracy of 44.5 achieved at step 600. In contrast, the second iteration improves performance to 45.0 (+0.5) for FP-target and 44.8 (+0.3) for FP-fixed over the original checkpoint. These results point to the possibility that iteratively refreshing failure prefixes may help recover additional learning signal and enable continued improvement.

\section{Related Work}

\textbf{Task Difficulty for RL training of LLMs.} Prior work has theoretically shown that policy updates in RL training of LLMs are biased by task difficulty, with the learning signal maximized when task success rates lie at an intermediate level \cite{li2025disco, razin2024vanishing, }. \citet{zhan2025exgrpo} further empirically validates this observation by stratifying training tasks by difficulty and showing that performance improvements peak on moderate-difficulty tasks. Building on this insight, several approaches regulate task difficulty during RL training. DAPO dynamically filters prompts with extremely low or high rollout accuracy that contribute little to training~\cite{yu2025dapo}. Some approaches arrange static training tasks in increasing order of difficulty to improve training efficiency \cite{chen2025selfevolving, kimi2025k1_5, shi2025efficient}. Instead of reordering a fixed task set, RLVE introduces an adaptive training environment that directly controls task difficulty relative to the model~\cite{zeng2025rlve}. Self-play approaches have also been explored, where LLMs autonomously propose tasks at an appropriate difficulty level~\cite{liu2025spiral, zhao2025absolutezero}. Our work shares the goal of maintaining effective task difficulty; however, rather than filtering or reordering tasks, we focus on saturated problems and recover their utility focusing on failure-prone states by conditioning on partial incorrect trajectories.

% \textbf{Task Difficulty for RL Training of LLMs.} Prior work shows that policy updates in RL training of LLMs depend strongly on task difficulty, with learning signals maximized when task success rates are at an intermediate level~\cite{li2025disco, razin2024vanishing}. \citet{zhan2025exgrpo} empirically supports this by stratifying tasks by difficulty and showing that gains peak on moderate-difficulty examples. Building on this insight, several methods regulate task difficulty during RL training. DAPO filters prompts with extremely low or high rollout accuracy, which contribute little to parameter updates~\cite{yu2025dapo}. Other approaches order fixed training tasks from easier to harder to improve training efficiency~\cite{chen2025selfevolving, kimi2025k1_5, shi2025efficient}. RLVE instead adapts the training environment to control task difficulty relative to the model~\cite{zeng2025rlve}. Self-play methods similarly let LLMs generate tasks at suitable difficulty levels~\cite{liu2025spiral, zhao2025absolutezero}. Our work shares the goal of maintaining effective difficulty, but differs by focusing on saturated problems: rather than filtering, reordering, or generating tasks, we recover their utility by conditioning training on failure-prone states from partial incorrect trajectories.

\textbf{Curriculum learning and context-enhanced learning for RL training of LLMs.}
Curriculum learning has been widely studied as a way to regulate task difficulty and improve training efficiency \cite{baker2020emergent,  jiang2020prioritized, portelas2020automatic}, and has recently been extended to RL training for large language models \cite{ chen2025selfevolving, gao2025promptcurriculum}. In the context of LLMs, many such approaches operate by augmenting the input context with additional information that facilitates reasoning during training \cite{xi2024reversecurriculum}. BREAD and POPE provide partial ground-truth hints in the context for difficult problems where the base model fails to produce successful rollouts, enabling the model to generate correct trajectories during training and yielding improvements that transfer effectively to settings without such hints \cite{qu2025pope, zhang2025bread}. Similarly, \citet{zhu2025contextenhanced} inject additional information into the context—excluded from autoregressive gradient computation—to achieve exponentially improved sample efficiency on multi-step reasoning tasks. Our approach also modifies the input context during RL training, but essentially inverts this paradigm: whereas prior methods inject \textit{correct} information to assist with \textit{difficult} problems, we condition on \textit{incorrect} trajectories to increase the difficulty of \textit{easy} (saturated) problems. This exposes failure-prone states and allows the recovery of learning signals that are otherwise poorly accessible.

% \textbf{Curriculum Learning and Context-Enhanced RL for LLMs.}
% Curriculum learning is widely used to regulate task difficulty and improve training efficiency~\cite{baker2020emergent, jiang2020prioritized, portelas2020automatic}, and has recently been applied to RL training for LLMs~\cite{chen2025selfevolving, gao2025promptcurriculum}. Several LLM methods further shape difficulty by augmenting the input context during training~\cite{xi2024reversecurriculum}. BREAD and POPE add partial ground-truth hints to help the model produce correct trajectories for difficult problems, with gains transferring to inference without hints~\cite{qu2025pope, zhang2025bread}. Similarly, \citet{zhu2025contextenhanced} inject auxiliary context, excluded from autoregressive gradient computation, to improve sample efficiency on multi-step reasoning tasks.
% Our method also modifies the training context, but inverts this paradigm. Rather than adding \textit{correct} information to make difficult problems easier, we add prefixes from \textit{incorrect} trajectories to make saturated problems informative again. This recovers learning signals that standard sampling rarely accesses.

\textbf{Scaling RL training for performance boosts.}
Scaling RL training has repeatedly been shown to improve the reasoning performance of LLMs \cite{olmo3_2025, guo2025deepseek}. ProRL demonstrates that substantially increasing the number of gradient updates can improve both pass@1 and pass@$k$ performance \cite{du2025prorl}. BroRL further extends this line of work by dramatically increasing per-problem exploration, sampling hundreds of rollouts per prompt to surpass the saturation point observed in ProRL and achieve additional gains \cite{hu2025brorl}. Relatedly, \citet{wang2025oneexample} show that intensive training on a single problem can extract sufficient learning signal to match the performance gains obtained from training on datasets with over a thousand examples. Taken together, these results indicate that scaling RL training through increased gradient steps or exploration can drive further performance gains. However, it incurs prohibitive computational costs and memory overhead, with the vast majority of the sampling budget being consumed by generating redundant correct solutions during training. In contrast, our approach extracts learning signals from saturated problems efficiently, reallocating exploration to failure-prone states rather than simply increasing the sampling budget.

% \textbf{Scaling RL training for performance boosts.}
% Scaling RL training has consistently improved LLM reasoning performance \cite{olmo3_2025, guo2025deepseek}. ProRL shows that increasing gradient updates can improve both pass@1 and pass@$k$ \cite{du2025prorl}, while BroRL further scales per-problem exploration by sampling hundreds of rollouts per prompt, surpassing ProRL’s saturation point and yielding additional gains \cite{hu2025brorl}. Similarly, \citet{wang2025oneexample} show that intensive training on a single problem can provide enough learning signal to match gains from datasets with over a thousand examples. Together, these results suggest that scaling RL through more updates or exploration can improve performance, but at prohibitive computational and memory cost: much of the sampling budget is spent generating redundant correct solutions. In contrast, our approach efficiently extracts learning signal from saturated problems by reallocating exploration toward failure-prone states rather than simply increasing the sampling budget.

\section{Conclusion}
In this paper, we introduce failure-prefix conditioning, a simple and effective method that extends RLVR training on saturated problems where standard methods stall. By explicitly initializing training from rare incorrect partial trajectories, our approach recovers critical learning signals, yielding performance gains comparable to training on medium-difficulty tasks while enhancing robustness to misleading reasoning. Moreover, we demonstrate that iteratively refreshing these prefixes enables sustained improvement beyond initial plateaus. Ultimately, our work establishes that saturated problems remain a valuable resource for training reasoning models, provided exploration is directed toward their sparse but informative failure modes.

\section{Limitations and Future Work}\label{sec:limitation}
While our work suggests that saturated problems can still be useful for RLVR with failure-prefix conditioning, it has several limitations that motivate future work.

First, failure-prefix conditioning requires collecting rare incorrect rollouts from saturated problems, which introduces considerable computational overhead (Appendix~\ref{appendix:overhead}). While existing work has largely focused on improving the probability of correct generation, an important future direction is to develop efficient methods for sampling rare incorrect on-policy trajectories from saturated problems.

Second, our experiments use a relatively small model, for which unsaturated training data remains accessible. However, failure-prefix conditioning is most relevant for frontier models, where acquiring new unsaturated problems becomes significantly more difficult. Further work is needed to determine whether the same benefits hold at larger scales.

Third, as discussed in Section~\ref{subsection:recovery}, failure-prefix conditioning improves robustness to misleading early reasoning steps, but may mildly reduce adherence to correct intermediate reasoning. Future work is needed to preserve recovery ability while maintaining consistency with correct reasoning.

Fourth, our experiments are limited to a single model and the math reasoning domain. Broader evaluation across model scales, model families, and task domains is needed to establish the generality of the approach.

\newpage
% \section*{References}
\bibliographystyle{plainnat}   % or unsrtnat, abbrvnat
\bibliography{ref}

@inproceedings{lambert2025tulu3,
  title        = {Tulu 3: Pushing Frontiers in Open Language Model Post-Training},
  author       = {Lambert, Nathan and Morrison, Jacob and Pyatkin, Valentina and Huang, Shengyi and Ivison, Hamish and Brahman, Faeze and Miranda, Lester James V. and Liu, Alisa and Dziri, Nouha and Lyu, Shane and Gu, Yuling and Malik, Saumya and Graf, Victoria and Hwang, Jena D. and Yang, Jiangjiang and Le Bras, Ronan and Tafjord, Oyvind and Wilhelm, Chris and Soldaini, Luca and Smith, Noah A. and Wang, Yizhong and Dasigi, Pradeep and Hajishirzi, Hannaneh},
  booktitle    = {Proceedings of the Conference on Language Models (COLM 2025)},
  year         = {2025},
}

@article{guo2025deepseek,
  title   = {DeepSeek-R1 incentivizes reasoning in LLMs through reinforcement learning},
  author  = {Guo, Daya and Yang, Dejian and Zhang, Haowei and Song, Junxiao and Wang, Peiyi and Zhu, Qihao and Xu, Runxin and Zhang, Ruoyu and Ma, Shirong and Bi, Xiao and Zhang, Xiaokang and Yu, Xingkai and Wu, Yu and Wu, Z. F. and Gou, Zhibin and Shao, Zhihong and Li, Zhuoshu and Gao, Ziyi and Liu, Aixin and ...},
  journal = {Nature},
  volume  = {645},
  number  = {8081},
  pages   = {633--638},
  year    = {2025},
  doi     = {10.1038/s41586-025-09422-z},
}

@misc{openai2024o1,
  title={OpenAI o1 System Card},
  author={OpenAI},
  year={2024},
  note={arXiv:2412.16720}
}

@inproceedings{razin2024vanishing,
  title        = {Vanishing Gradients in Reinforcement Finetuning of Language Models},
  author       = {Razin, Noam and Zhou, Hattie and Saremi, Omid and Thilak, Vimal and Bradley, Arwen and Nakkiran, Preetum and Susskind, Joshua M. and Littwin, Etai},
  booktitle    = {Proceedings of International Conference on Learning Representations (ICLR 2024)},
  year         = {2024}
}

@inproceedings{setlur2025e3,
  title   = {e3: Learning to Explore Enables Extrapolation of Test-Time Compute for LLMs},
  author  = {Setlur, Amrith and Yang, Matthew Y. R. and Snell, Charlie Victor and Greer, Jeremiah and Wu, Ian and Smith, Virginia and Simchowitz, Max and Kumar, Aviral},
  booktitle = {Proceedings of the International Conference on Learning Representations 2026 (ICLR 2026)},
  year      = {2026},
  note      = {arXiv:2506.09026},
}

@article{Kim2025RLvsDistill,
  title   = {Reinforcement Learning vs. Distillation: Understanding Accuracy and Capability in LLM Reasoning},
  author  = {Kim, Minwu and Shrestha, Anubhav and Shrestha, Safal and Nepal, Aadim and Ross, Keith},
  journal = {arXiv:2505.14216},
  year    = {2025},
}

@article{hu2025brorl,
  title   = {BroRL: Scaling Reinforcement Learning via Broadened Exploration},
  author  = {Hu, Jian and Liu, Mingjie and Lu, Ximing and Wu, Fang and Harchaoui, Zaid and Diao, Shizhe and Choi, Yejin and Molchanov, Pavlo and Yang, Jun and Kautz, Jan and Dong, Yi},
  journal = {arXiv:2510.01180},
  year    = {2025},
}

@inproceedings{zhu2025negative,
  title     = {The Surprising Effectiveness of Negative Reinforcement in LLM Reasoning},
  author    = {Zhu, Xinyu and Xia, Mengzhou and Wei, Zhepei and Chen, Wei-Lin and Chen, Danqi and Meng, Yu},
  booktitle = {Proceedings of the Thirty-Ninth Conference on Neural Information Processing Systems (NeurIPS 2025)},
  year      = {2025},
  note      = {arXiv:2506.01347},
}

@inproceedings{wang2025oneexample,
  title     = {Reinforcement Learning for Reasoning in Large Language Models with One Training Example},
  author    = {Wang, Yiping and Yang, Qing and Zeng, Zhiyuan and Ren, Liliang and Liu, Lucas and Peng, Baolin and Cheng, Hao and He, Xuehai and Wang, Kuan and Gao, Jianfeng and Chen, Weizhu and Wang, Shuohang and Du, Simon Shaolei and Shen, Yelong},
  booktitle = {Proceedings of the Thirty-Ninth Conference on Neural Information Processing Systems (NeurIPS 2025)},
  year      = {2025},
  note      = {arXiv:2504.20571},
}

@inproceedings{yu2025dapo,
  title     = {DAPO: An Open-Source LLM Reinforcement Learning System at Scale},
  author    = {Yu, Qiying and Zhang, Zheng and Zhu, Ruofei and Yuan, Yufeng and Zuo, Xiaochen and Yue, Yu and Fan, Tiantian and Liu, Gaohong and Liu, Lingjun and Liu, Xin and Lin, Haibin and Lin, Zhiqi and Ma, Bole and Sheng, Guangming and Tong, Yuxuan and Zhang, Chi and Zhang, Mofan and Zhang, Wang and Zhu, Hang and Zhu, Jinhua and Chen, Jiaze and Chen, Jiangjie and Wang, Chengyi and Yu, Hongli and Song, Yuxuan and Wei, Xiangpeng and Zhou, Hao and Liu, Jingjing and Ma, Wei-Ying and Zhang, Ya-Qin and Yan, Lin and Qiao, Mu and Wu, Yonghui and Wang, Mingxuan},
  booktitle = {Proceedings of the Thirty-Ninth Conference on Neural Information Processing Systems (NeurIPS 2025)},
  year      = {2025},
  note      = {arXiv:2503.14476},
}

@inproceedings{li2025disco,
  title     = {DisCO: Reinforcing Large Reasoning Models with Discriminative Constrained Optimization},
  author    = {Li, Gang and Lin, Ming C. and Galanti, Tomer and Tu, Zhengzhong and Yang, Tianbao},
  booktitle = {Proceedings of the Thirty-Ninth Conference on Neural Information Processing Systems (NeurIPS 2025)},
  year      = {2025},
  note      = {arXiv:2505.12366},
}

@inproceedings{zeng2025rlve,
  title   = {RLVE: Scaling Up Reinforcement Learning for Language Models with Adaptive Verifiable Environments},
  author  = {Zeng, Zhiyuan and Ivison, Hamish and Wang, Yiping and Yuan, Lifan and Li, Shuyue Stella and Ye, Zhuorui and Li, Siting and He, Jacqueline and Zhou, Runlong and Chen, Tong and Zhao, Chenyang and Tsvetkov, Yulia and Du, Simon Shaolei and Jaques, Natasha and Peng, Hao and Koh, Pang Wei and Hajishirzi, Hannaneh},
  booktitle = {Proceedings of the 43rd International Conference on Machine Learning Representations 2026 (ICML 2026)},
  year      = {2026},
  note      = {arXiv:2511.07317},
}

@article{kimi2025k1_5,
  title   = {Kimi k1.5: Scaling Reinforcement Learning with LLMs},
  author  = {Du, Angang and Gao, Bofei and Xing, Bowei and Jiang, Changjiu and Chen, Cheng and Li, Cheng and Xiao, Chenjun and Du, Chenzhuang and Liao, Chonghua and Tang, Chuning and Wang, Congcong and Zhang, Dehao and Yuan, Enming and Lu, Enzhe and Tang, Fengxiang and Sung, Flood and Wei, Guangda and Lai, Guokun and ...},
  journal = {arXiv:2501.12599},
  year    = {2025},
}

@article{shi2025efficient,
  title   = {Efficient Reinforcement Finetuning via Adaptive Curriculum Learning},
  author  = {Shi, Taiwei and Wu, Yiyang and Song, Linxin and Zhou, Tianyi and Zhao, Jieyu},
  journal = {arXiv:2504.05520},
  year    = {2025},
}

@article{chen2025selfevolving,
  title   = {Self-Evolving Curriculum for LLM Reasoning},
  author  = {Chen, Xiaoyin and Lu, Jiarui and Kim, Minsu and Zhang, Dinghuai and Tang, Jian and Piché, Alexandre and Gontier, Nicolas and Bengio, Yoshua and Kamalloo, Ehsan},
  journal = {arXiv:2505.14970},
  year    = {2025},
}

@inproceedings{zhao2025absolutezero,
  title     = {Absolute Zero: Reinforced Self-Play Reasoning with Zero Data},
  author    = {Zhao, Andrew and Wu, Yiran and Yue, Yang and Wu, Tong and Xu, Quentin and Lin, Matthieu and Wang, Shenzhi and Wu, Qingyun and Zheng, Zilong and Huang, Gao},
  booktitle = {Proceedings of the Thirty-Ninth Conference on Neural Information Processing Systems (NeurIPS 2025)},
  year      = {2025},
  note      = {arXiv:2505.03335},
}

@inproceedings{liu2025spiral,
  title     = {SPIRAL: Self-Play on Zero-Sum Games Incentivizes Reasoning via Multi-Agent Multi-Turn Reinforcement Learning},
  author    = {Liu, Bo and Guertler, Leon and Yu, Simon and Liu, Zichen and Qi, Penghui and Balcells, Daniel and Liu, Mickel and Tan, Cheston and Shi, Weiyan and Lin, Min and Lee, Wee Sun and Jaques, Natasha},
  booktitle = {Proceedings of the International Conference on Learning Representations 2026 (ICLR 2026)},
  year      = {2025},
  note      = {arXiv:2506.24119},
}

@article{gao2025promptcurriculum,
  title   = {Prompt Curriculum Learning for Efficient LLM Post-Training},
  author  = {Gao, Zhaolin and Kim, Joongwon and Sun, Wen and Joachims, Thorsten and Wang, Sid and Pang, Richard Yuanzhe and Tan, Liang},
  journal = {arXiv:2510.01135},
  year    = {2025},
}

@article{qu2025pope,
  title   = {POPE: Learning to Reason on Hard Problems via Privileged On-Policy Exploration},
  author={Qu, Yuxiao and Setlur, Amrith and Smith, Virginia and Salakhutdinov, Ruslan and Kumar, Aviral},
  journal = {arXiv:2601.18779},
  year    = {2026},
}

@inproceedings{baker2020emergent,
  title     = {Emergent Tool Use From Multi-Agent Autocurricula},
  author    = {Baker, Bowen and Kanitscheider, Ingmar and Markov, Todor M. and Wu, Yi and Powell, Glenn and McGrew, Bob and Mordatch, Igor},
  booktitle = {Proceedings of the International Conference on Learning Representations (ICLR 2020)},
  year      = {2020},
}

@inproceedings{portelas2020automatic,
  title     = {Automatic Curriculum Learning for Deep Reinforcement Learning: A Short Survey},
  author    = {Portelas, Rapha{\"e}l and Colas, Cl{\'e}ment and Weng, Lucas and Hofmann, Karl and Oudeyer, Pierre-Yves},
  booktitle = {Proceedings of the 29th International Joint Conference on Artificial Intelligence (IJCAI 2020)},
  year      = {2020}
}

@inproceedings{jiang2020prioritized,
  title     = {Prioritized Level Replay},
  author    = {Jiang, Mingxuan and Grefenstette, Edward and Rockt{\"a}schel, Tim},
  booktitle = {Proceedings of the 37th International Conference on Machine Learning (ICML 2020)},
  year      = {2020}
}

@inproceedings{zhang2025bread,
  title     = {BREAD: Branched Rollouts from Expert Anchors Bridge SFT \& RL for Reasoning},
  author    = {Zhang, Xuechen and Huang, Zijian and Li, Yingcong and Ni, Chenshun and Chen, Jiasi and Oymak, Samet},
  booktitle = {Proceedings of the Thirty-Ninth Conference on Neural Information Processing Systems (NeurIPS 2025)},
  year      = {2025},
  note      = {arXiv:2506.17211},
}

@inproceedings{xi2024reversecurriculum,
  title     = {Training Large Language Models for Reasoning through Reverse Curriculum Reinforcement Learning},
  author    = {Xi, Zhiheng and Chen, Wenxiang and Hong, Boyang and Jin, Senjie and Zheng, Rui and He, Wei and Ding, Yiwen and Liu, Shichun and Guo, Xin and Wang, Junzhe and Guo, Honglin and Shen, Wei and Fan, Xiaoran and Zhou, Yuhao and Dou, Shihan and Wang, Xiao and Zhang, Xinbo and Sun, Peng and Gui, Tao and Zhang, Qi and Huang, Xuanjing},
  booktitle = {Proceedings of the 41st International Conference on Machine Learning (ICML 2024)},
  year      = {2024},
}

@inproceedings{zhu2025contextenhanced,
  title     = {On the Power of Context-Enhanced Learning in LLMs},
  author    = {Zhu, Xingyu and Panigrahi, Abhishek and Arora, Sanjeev},
  booktitle = {Proceedings of the 42nd International Conference on Machine Learning (ICML 2025)},
  year      = {2025},
  note      = {arXiv:2503.01821},
}

@inproceedings{du2025prorl,
  title     = {ProRL: Prolonged Reinforcement Learning Expands Reasoning Strategies},
  author    = {Du, Angang and Gao, Bofei and Xing, Bowei and Jiang, Changjiu and Chen, Cheng and Li, Cheng and Xiao, Chenjun and Du, Chenzhuang and Liao, Chonghua and ...},
  booktitle = {Proceedings of the Thirty-Ninth Conference on Neural Information Processing Systems (NeurIPS 2025)},
  year      = {2025},
  note      = {arXiv:2505.24864},
}

@article{olmo3_2025,
  title   = {Olmo 3},
  author  = {Ettinger, Allyson and Bertsch, Amanda and Kuehl, Bailey and Graham, David and Heineman, David and Groeneveld, Dirk and Brahman, Faeze and Timbers, Finbarr and Ivison, Hamish and Morrison, Jacob and Poznanski, Jake and Lo, Kyle and Soldaini, Luca and Jordan, Matt and Chen, Mayee and Noukhovitch, Michael and Lambert, Nathan and Walsh, Pete and Dasigi, Pradeep and Berry, Robert and Malik, Saumya and Shah, Saurabh and Geng, Scott and Arora, Shane and Gupta, Shashank and Anderson, Taira and Xiao, Teng and Murray, Tyler and Romero, Tyler and Graf, Victoria and Asai, Akari and Bhagia, Akshita and Wettig, Alexander and Liu, Alisa and Rangapur, Aman and Anastasiades, Chloe and Huang, Costa and Schwenk, Dustin and Trivedi, Harsh and Magnusson, Ian and Lochner, Jaron and Liu, Jiacheng and Miranda, Lester James V. and Sap, Maarten and Morgan, Malia and Schmitz, Michael and Guerquin, Michal and Wilson, Michael and Huff, Regan and Le Bras, Ronan and Xin, Rui and Shao, Rulin and Skjonsberg, Sam and Shen, Shannon Zejiang and Li, Shuyue Stella and Wilde, Tucker and Pyatkin, Valentina and Merrill, Will and Chang, Yapei and Gu, Yuling and Zeng, Zhiyuan and Sabharwal, Ashish and Zettlemoyer, Luke and Koh, Pang Wei and Farhadi, Ali and Smith, Noah A. and Hajishirzi, Hannaneh},
  journal = {arXiv:2512.13961},
  year    = {2025},
}

@misc{shao2024deepseekmath,
  title={DeepSeekMath: Pushing the Limits of Mathematical Reasoning in Open Language Models},
  author={Zhihong Shao and Peiyi Wang and Qihao Zhu and Runxin Xu and Junxiao Song and Xiao Bi and Haowei Zhang and Mingchuan Zhang and Y. K. Li and Y. Wu and Daya Guo},
  year={2024},
  primaryClass={cs.CL},
  note      = {arXiv:2402.03300},

}

@article{zhan2025exgrpo,
  title     = {ExGRPO: Learning to Reason from Experience},
  author    = {Runzhe Zhan and Yafu Li and Zhi Wang and Xiaoye Qu and Dongrui Liu and Jing Shao and Derek F. Wong and Yu Cheng},
  year      = {2025},
  note      = {arXiv:2510.02245},

}

@inproceedings{hendrycks2021measuring,
  title        = {Measuring Mathematical Problem Solving With the MATH Dataset},
  author       = {Hendrycks, Dan and Burns, Collin and Kadavath, Saurav and Arora, Akul and Basart, Steven and Tang, Eric and Song, Dawn and Steinhardt, Jacob},
  booktitle    = {Proceedings of the Thirty-Fifth Conference on Neural Information Processing Systems (NeurIPS 2021) Track on Datasets and Benchmarks},
  year         = {2021},

}

@misc{aime24dataset,
  title        = {AIME24 Dataset},
  author       = {{math-ai}},
  howpublished = {\url{https://huggingface.co/datasets/math-ai/aime24}},
  year         = {2025},
  note         = {Hugging Face dataset},
}

@misc{aime25dataset,
  title        = {AIME25 Dataset},
  author       = {{math-ai}},
  howpublished = {\url{https://huggingface.co/datasets/math-ai/aime25}},
  year         = {2025},
  note         = {Hugging Face dataset},
}

@misc{amc12dataset,
  title        = {AMC12 2022, 2023 Dataset},
  author       = {{AI-MO}},
  howpublished = {\url{https://huggingface.co/datasets/AI-MO/aimo-validation-amc}},
  year         = {2025},
  note         = {Hugging Face dataset},
}

@misc{hmmt25dataset,
  title        = {HMMT25 Dataset (February 2025)},
  author       = {{MathArena}},
  howpublished = {\url{https://huggingface.co/datasets/MathArena/hmmt_feb_2025}},
  year         = {2025},
  note         = {Hugging Face dataset},
}

@article{wen2025parathinker,
  title        = {ParaThinker: Native Parallel Thinking as a New Paradigm to Scale LLM Test-time Compute},
  author       = {Hao Wen and Yifan Su and Feifei Zhang and Yunxin Liu and Yunhao Liu and Ya-Qin Zhang and Yuanchun Li},
  year         = {2025},
  note      = {arXiv:2510.02245},
}

@inproceedings{kwon2023efficient,
  title        = {Efficient Memory Management for Large Language Model Serving with PagedAttention},
  author       = {Kwon, Woosuk and Li, Zhuohan and Zhuang, Siyuan and Sheng, Ying and Zheng, Lianmin and Yu, Cody Hao and Gonzalez, Joseph and Zhang, Hao and Stoica, Ion},
  booktitle    = {Proceedings of the 29th {ACM} Symposium on Operating Systems Principles (SOSP ’23)},
  year         = {2023},
  pages        = {611--626},
  publisher    = {Association for Computing Machinery},
  address      = {Koblenz, Germany},
  doi          = {10.1145/3600006.3613165}
}

@inproceedings{deepScaler2025,
  title        = {DeepScaleR: Effective RL Scaling of Reasoning Models via Iterative Context Lengthening},
  author       = {Anonymous},
  booktitle    = {ICLR 2026 Conference Submission},
  year         = {2025},
  url          = {https://openreview.net/forum?id=I6GzDCne7U},
  note         = {Under review}
}

@misc{novikov2025alphaevolve,
  title  = {AlphaEvolve: A coding agent for scientific and algorithmic discovery},
  author = {Novikov, Alexander and V{\~u}, Ng{\^a}n and Eisenberger, Marvin and Dupont, Emilien and Huang, Po-Sen and Wagner, Adam Zsolt and Shirobokov, Sergey and Kozlovskii, Borislav and Ruiz, Francisco J. R. and Mehrabian, Abbas and Kumar, M. Pawan and See, Abigail and Chaudhuri, Swarat and Holland, George and Davies, Alex and Nowozin, Sebastian and Kohli, Pushmeet and Balog, Matej},
  year   = {2025},
  note   = {arXiv:2506.13131},
}

@misc{feng2026gemini_erdos,
  title  = {Semi-Autonomous Mathematics Discovery with Gemini: A Case Study on the Erd{\H{o}}s Problems},
  author = {Feng, Tony and Trinh, Trieu and Bingham, Garrett and Kang, Jiwon and Zhang, Shengtong and Kim, Sang-hyun and Barreto, Kevin and Schildkraut, Carl and Jung, Junehyuk and Seo, Jaehyeon and Pagano, Carlo and Chervonyi, Yuri and Hwang, Dawsen and Hou, Kaiying and Gukov, Sergei and Tsai, Cheng-Chiang and Choi, Hyunwoo and Jin, Youngbeom and Li, Wei-Yuan and Wu, Hao-An and Shiu, Ruey-An and Shih, Yu-Sheng and Le, Quoc V. and Luong, Thang},
  year   = {2026},
  note   = {arXiv:2601.22401},
}

@misc{alexeev2026shortproofs2,
  title  = {Short proofs in combinatorics, probability and number theory II},
  author = {Alexeev, Boris and Putterman, Moe and Sawhney, Mehtaab and Sellke, Mark and Valiant, Gregory},
  year   = {2026},
  note   = {arXiv:2604.06609},
}

@misc{akhtar2026benchmark_saturation,
  title  = {When AI Benchmarks Plateau: A Systematic Study of Benchmark Saturation},
  author = {Akhtar, Mubashara and Reuel, Anka and Soni, Prajna and Ahuja, Sanchit and Ammanamanchi, Pawan Sasanka and Rawal, Ruchit and Zouhar, Vil{\'e}m and Yadav, Srishti and Whitehouse, Chenxi and Ki, Dayeon and Mickel, Jennifer and Choshen, Leshem and {\v{S}}uppa, Marek and Batzner, Jan and Chim, Jenny and Sania, Jeba and Long, Yanan and Rahmani, Hossein A. and Knight, Christina and Nan, Yiyang and Raj, Jyoutir and Fan, Yu and Singh, Shubham and Sahoo, Subramanyam and Habba, Eliya and Gohar, Usman and Pawar, Siddhesh and Scholz, Robert and Subramonian, Arjun and Ni, Jingwei and Kochenderfer, Mykel and Koyejo, Sanmi and Sachan, Mrinmaya and Biderman, Stella and Talat, Zeerak and Ghosh, Avijit and Solaiman, Irene},
  year   = {2026},
  note   = {arXiv:2602.16763},
}

@inproceedings{luong2025robust_math_reasoning,
  title={Towards Robust Mathematical Reasoning},
  author={Luong, Thang and Hwang, Dawsen and Nguyen, Hoang H. and Ghiasi, Golnaz and Chervonyi, Yuri and Seo, Insuk and Kim, Junsu and Bingham, Garrett and Lee, Jonathan and Mishra, Swaroop and Zhai, Alex and Hu, Clara Huiyi and Michalewski, Henryk and Kim, Jimin and Ahn, Jeonghyun and Bae, Junhwi and Song, Xingyou and Trinh, Trieu H. and Le, Quoc V. and Jung, Junehyuk},
  booktitle={Proceedings of the 2025 Conference on Empirical Methods in Natural Language Processing (EMNLP 2025)},
  year={2025}
}

@misc{openai2025competitive_programming,
  title  = {Competitive Programming with Large Reasoning Models},
  author = {El-Kishky, Ahmed and Wei, Alexander and Saraiva, Andre and Minaiev, Borys and Selsam, Daniel and Dohan, David and Song, Francis and Lightman, Hunter and Clavera, Ignasi and Pachocki, Jakub and Tworek, Jerry and Kuhn, Lorenz and Kaiser, Lukasz and Chen, Mark and Schwarzer, Max and Rohaninejad, Mostafa and McAleese, Nat and others},
  year   = {2025},
  note   = {arXiv:2502.06807},
}

@misc{deepmind2025gemini_icpc,
  title  = {Gemini achieves gold-medal-level at the International Collegiate Programming Contest World Finals},
  author = {{Google DeepMind}},
  howpublished = {\url{https://deepmind.google/blog/gemini-achieves-gold-medal-level-at-the-international-collegiate-programming-contest-world-finals/}},
  year   = {2025},
}

@misc{villalobos2024run_out_of_data,
  title  = {Will we run out of data? Limits of LLM scaling based on human-generated data},
  author = {Villalobos, Pablo and Ho, Anson and Sevilla, Jaime and Besiroglu, Tamay and Heim, Lennart and Hobbhahn, Marius},
  year   = {2024},
  note   = {arXiv:2211.04325},
}

@inproceedings{liu2025r1zero,
  title={Understanding R1-Zero-Like Training: A Critical Perspective},
  author={Liu, Zichen and Chen, Changyu and Li, Wenjun and Qi, Penghui and Pang, Tianyu and Du, Chao and Lee, Wee Sun and Lin, Min},
  booktitle={Proceedings of the 2025 Conference on Language Modeling (COLM 2025)},
  year={2025}
}

%%%%%%%%%%%%%%%%%%%%%%%%%%%%%%%%%%%%%%%%%%%%%%%%%%%%%%%%%%%%
\newpage

\appendix

\section{Failure-Prefix Conditioning Algorithms}
\label{appendix:failure-prefix-algorithms}

Algorithm~\ref{alg:failure-prefix-conditioning} describes the full failure-prefix conditioning procedure used in our main experiments. For each saturated question, we first obtain an incorrect rollout and construct multiple candidate prefixes by truncating it at different lengths. We then estimate the rollout accuracy induced by each prefix and select the one whose accuracy is closest to the target accuracy \(\tau\). This produces a prefix-conditioned dataset that shifts exploration toward failure-prone states with high reward variance.

\begin{algorithm}
\caption{Failure-Prefix Conditioning with Target Accuracy}
\label{alg:failure-prefix-conditioning}
\begin{algorithmic}[1]
\Require Saturated dataset $\mathcal{D}_{\text{saturated}} = \{(q, a^\ast)\}$, policy $\pi_\theta$, target accuracy $\tau$, number of rollouts $N$
\Ensure Prefix-conditioned dataset $\mathcal{D}'$

\State Initialize $\mathcal{D}' \leftarrow \emptyset$
\For{each $(q, a^\ast) \in \mathcal{D}_{\text{saturated}}$}
    \State Obtain an incorrect rollout $\tilde{y} \sim \pi_\theta(\cdot \mid q)$
    \State Construct candidate prefix set
    \[
        \mathcal{S}(q) = \{\tilde{y}_{1:\alpha_k}\}_{k=1}^K
    \]
    \For{each prefix $s \in \mathcal{S}(q)$}
        \State Sample $N$ rollouts $\{y_i\}_{i=1}^N \sim \pi_\theta(\cdot \mid q \oplus s)$
        \State Estimate prefix-conditioned rollout accuracy
        \[
        p_\theta(q,s) \leftarrow \frac{1}{N}\sum_{i=1}^N \mathbb{I}[r(y_i;q)=1]
        \]
    \EndFor
    \State Select prefix
    \[
    s^{(q)} \leftarrow \arg\min_{s \in \mathcal{S}(q)}
    \left|p_\theta(q,s) - \tau \right|
    \]
    \State Add $(q \oplus s^{(q)}, a^\ast)$ to $\mathcal{D}'$
\EndFor
\State \Return $\mathcal{D}'$
\end{algorithmic}
\end{algorithm}

Algorithm~\ref{alg:fixed-truncation} gives the lower-cost fixed-truncation variant. Instead of estimating rollout accuracy for multiple candidate prefixes, it keeps a fixed fraction \(\gamma\) of each incorrect rollout. This avoids the prefix-selection overhead while preserving most of the benefit of failure-prefix conditioning.

\begin{algorithm}
\caption{Failure-Prefix Conditioning with Fixed Truncation}
\label{alg:fixed-truncation}
\begin{algorithmic}[1]
\Require Saturated dataset $\mathcal{D}_{\text{saturated}} = \{(q, a^\ast)\}$, policy $\pi_\theta$, truncation fraction $\gamma \in (0,1)$
\Ensure Prefix-conditioned dataset $\mathcal{D}'_{\gamma}$

\State Initialize $\mathcal{D}' \leftarrow \emptyset$
\For{each $(q, a^\ast) \in \mathcal{D}_{\text{saturated}}$}
    \State Obtain an incorrect rollout $\tilde{y} \sim \pi_\theta(\cdot \mid q)$
    \State Let $L \leftarrow |\tilde{y}|$
    \State Set prefix length $\alpha \leftarrow \lfloor \gamma L \rfloor$
    \State Construct fixed-truncation prefix $s^{(q)} \leftarrow \tilde{y}_{1:\alpha}$
    \State Add $(q \oplus s^{(q)}, a^\ast)$ to $\mathcal{D}'$
\EndFor
\State \Return $\mathcal{D}'$
\end{algorithmic}
\end{algorithm}

\newpage

\section{GRPO Training Details}\label{appendix:GRPO_training}

\paragraph{Library}
For GRPO training in this work, we use the \texttt{GRPOTrainer} from the \texttt{TRL\footnote{\url{https://github.com/huggingface/trl}}} library

\paragraph{Prompt Setting}
We use the Qwen-Math chat template for all GRPO training and evaluation. When applying failure-prefix conditioning, the prefix is inserted immediately after the \texttt{<think>} token, as shown below.

\begin{tcolorbox}[title=Qwen-Math template, colback=blue!10, colframe=blue!50, fonttitle=\bfseries]

\footnotesize
\begin{verbatim}
<|im_start|>system
Please reason step by step, and put your final answer within \boxed{}.
<|im_end|>
<|im_start|>user
{question}
<|im_end|>
<|im_start|>assistant
<think>{prefix}
\end{verbatim}
\end{tcolorbox}

\paragraph{Reward Function}

We adopt a minimalistic reward setting. A response received a reward of 1 if it contained the correct final answer, and 0 otherwise. Answer verification is performed using the \texttt{math\_verify}\footnote{\url{https://github.com/huggingface/Math-Verify}} package, consistent with the one used for evaluation.

\[
R(q, a^*, y) =
\begin{cases}
1 & \text{if the response $y$ to question $q$ matches the ground truth answer $a^*$} \\
0 & \text{otherwise}
\end{cases}
\]

\paragraph{Clip-Higher}
Following DAPO~\cite{yu2025dapo}, we adopt the clip-higher strategy to mitigate entropy collapse during training. Specifically, we set $\epsilon_{\text{high}} = 0.4$ and $\epsilon_{\text{low}} = 0.2$, with number of iterations per batch as 2.

\paragraph{Training Hyperparameters}\label{appendix:GRPO_hyperparams}
Table~\ref{tab:grpo-hyperparams} summarizes the key hyperparameters used in RLVR training. Note that, with this configuration, running approximately 1,200 gradient steps takes about 5 days on 4 A100 80GB GPUs, corresponding to roughly 480 GPU-hours.
\begin{table}[H]
\centering
\caption{Key hyperparameters used for RLVR training.}
\begin{tabularx}{\linewidth}{@{}lX@{}}

\toprule
\textbf{Hyperparameter} & \textbf{Value} \\
\midrule
Optimizer & AdamW \\
Learning rate scheduler & Constant \\
Maximum response length & 8000 \\
Temperature & 1.0 \\
Top-$p$ & 1.0 \\
Number of rollouts per question & 16 \\
Number of iterations per batch  &  2 \\
$\epsilon_{\text{high}}$ &  0.4 \\
$\epsilon_{\text{hlow}}$ &  0.2 \\ 

Global batch size & 4 (per device) $\times$ 4 (GPUs) $\times$ 10 (accumulation) = 160 \\
Learning rate & $1 \times 10^{-6}$ \\
Tensor parallel size & 2\\
KL divergence weight & 0 \\

\bottomrule
\end{tabularx}

\label{tab:grpo-hyperparams}
\end{table}

\newpage

\section{Inference Details}
\label{appendix:inference_details}

\paragraph{Libraries}
For inference, we use \texttt{vLLM}~\cite{kwon2023efficient} for response generation. For response grading, we use the \texttt{math\_verify} package, consistent with the one used for rewarding during RLVR training.

\paragraph{Prompt Setting} For all evaluation and rollout accuracy measurement experiments, we use the same prompt setting described in Appendix~\ref{appendix:GRPO_training}.

\paragraph{Inference Hyperparameters for Measuring Rollout Accuracy}
To measure rollout accuracy for identifying saturated questions and determining the best prefix length, we run inference from each prefix-conditioned question using the same setting as in generating rollouts during GRPO training (Appendix~\ref{appendix:GRPO_hyperparams}). Specifically, we use the configuration shown in Table~\ref{tab:prefix_inference_details}. Note that, when identifying saturated questions, If the single incorrect rollout exceeds the maximum generation length of 8000 tokens, we exclude the corresponding question from the dataset, since responses truncated at the token limit cannot be reliably classified as correct or incorrect. In practice, this exclusion affects very few questions: problems with rollout accuracy $\geq 120/128$ are generally not difficult, and almost all generated responses terminate well within the 8000-token limit.

\begin{table}[H]
\caption{Key hyperparameters used for inference for identifying saturated questions and determining best prefix length.}
\centering
\begin{tabularx}{\linewidth}{@{}lX@{}}
\toprule
\textbf{Hyperparameter} & \textbf{Value} \\
\midrule
Maximum token length & 8000 \\
Temperature & 1.0 \\
Top-$p$ & 1.0 \\
Number of rollouts per question & 32 \\

\bottomrule
\end{tabularx}
\label{tab:prefix_inference_details}
\end{table}

\paragraph{Inference Hyperparameters for Evaluation}
Table~\ref{tab:eval_detail} summarizes the key hyperparameters used in inference for evaluation in Section~\ref{sec:main_results}, Section~\ref{sec:fixed_truncation} and Section~\ref{sec:round2}. The same setting is also used for measuring the recovery from failure prefixes in Section~\ref{subsection:recovery}.

\begin{table}[H]
\centering
\caption{Key hyperparameters used for inference during evaluation.}
\begin{tabularx}{\linewidth}{@{}lX@{}}
\toprule
\textbf{Hyperparameter} & \textbf{Value} \\
\midrule
Maximum token length & 32000 \\
Temperature & 0.6 \\
Top-$p$ & 1.0 \\
Number of rollouts per question & 32 \\

\bottomrule
\end{tabularx}
\label{tab:eval_detail}
\end{table}

\newpage

\section{Training Data Details}\label{appendix:train-set}

\begin{figure}[h]
    \centering
    \includegraphics[width=0.35\linewidth]{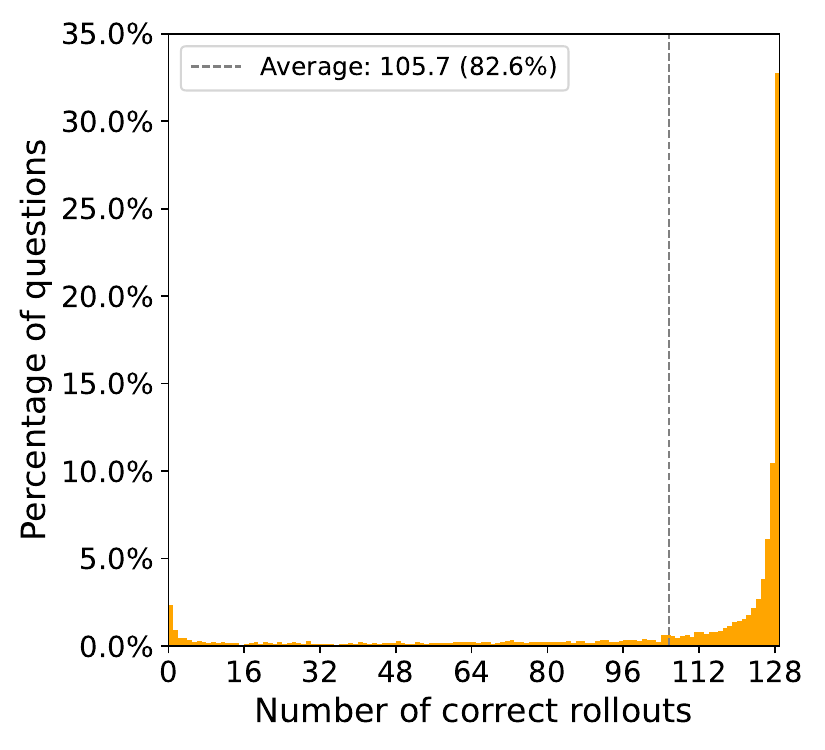}
    \vspace{-10pt}
    \caption{Question-level rollout accuracy of the base model on the MATH training set.}
    \label{fig:rollout-accuracy}
\end{figure}

\paragraph{MATH Train Set.}
Figure~\ref{fig:rollout-accuracy} shows the distribution of question-level rollout accuracy for the base model on the MATH training set.

\begin{figure}[h]
    \centering
    \includegraphics[width=0.95\linewidth]{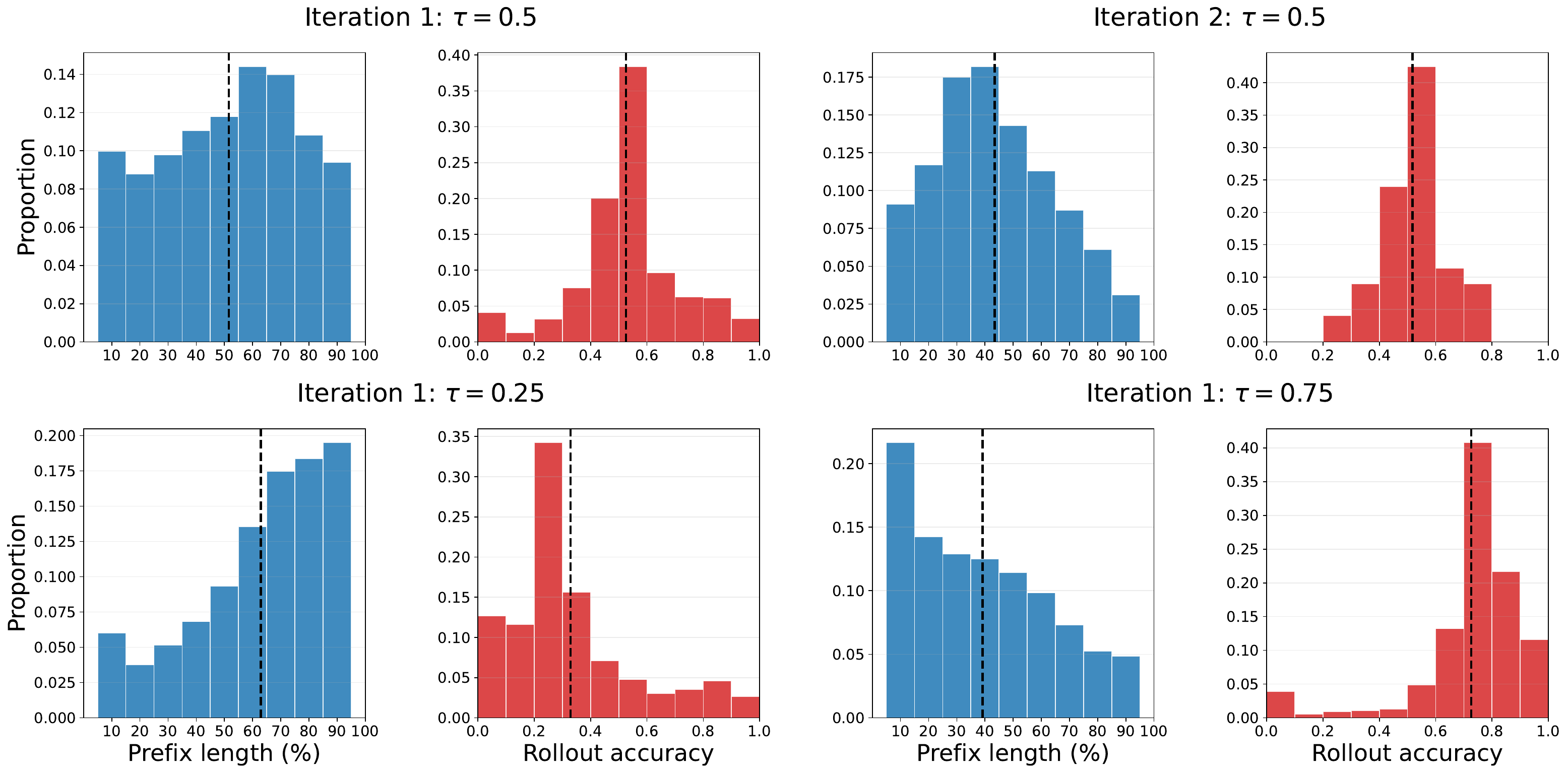}
    \vspace{-10pt}
    \caption{Failure-prefix-conditioned datasets selected by target accuracy. Each pair of plots shows the distribution of selected prefix lengths and the resulting prefix-conditioned rollout accuracies. Dashed lines indicate mean values.}
    \label{fig:failure-prefix-dataset_target}
\end{figure}

\begin{figure}[h]
    \centering
    \includegraphics[width=0.92\linewidth]{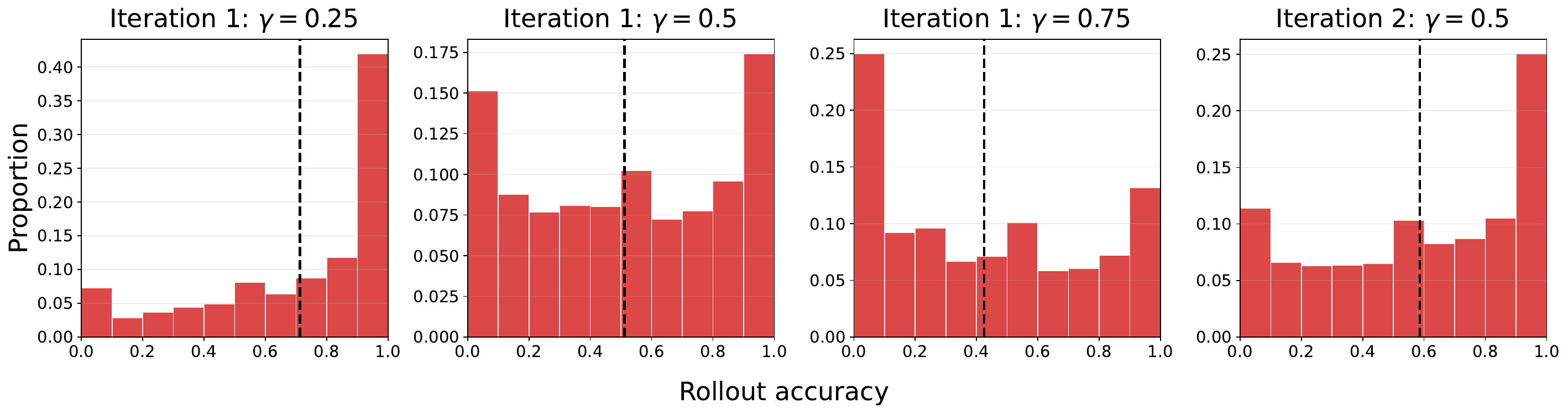}
    \vspace{-13pt}
    \caption{Prefix-conditioned rollout accuracy distributions for fixed-truncation datasets with different \(\gamma\) values. Dashed lines indicate mean values.}
    \label{fig:failure-prefix-dataset_FT}
\end{figure}

\paragraph{Failure-Prefix-Conditioned Dataset}
We construct failure-prefix-conditioned datasets using two selection strategies: target accuracy and fixed truncation. Figure~\ref{fig:failure-prefix-dataset_target} shows the distributions of selected prefix lengths and their resulting rollout accuracies under different target values \(\tau\), covering both the main experiments in Section~\ref{sec:main_results} and iterative failure-prefix conditioning in Section~\ref{sec:round2}. Figure~\ref{fig:failure-prefix-dataset_FT} reports the corresponding rollout-accuracy distributions for fixed truncation with different \(\gamma\) values.

\newpage 

\section{Computation Overhead for Failure-Prefix Conditioning}
\label{appendix:overhead}

\begin{table}[h]
\centering
\small
\begin{tabular}{lccc}
\toprule
Stage & Full cost & Reduced cost & Relative cost \\
\midrule
Saturation identification 
& 3.32B tokens 
& 0.64B tokens 
& 19.3\% \\
Prefix selection 
& 1.94B tokens 
& 0.11B tokens 
& 5.8\% \\
\bottomrule
\end{tabular}
\vspace{1em}
\caption{
Token overhead of failure-prefix conditioning and simple cost-reduction strategies. 
For saturation identification, staged sampling reduces the cost from 128 rollouts per question to a selective procedure that continues sampling only for initially saturated questions. 
For prefix selection, using a coarser 20\% prefix grid and 8 rollouts per prefix substantially reduces the estimation cost.
}
\label{tab:overhead}
\end{table}

Failure-prefix conditioning introduces additional computation in two stages. First, we sample multiple responses per training question to estimate rollout accuracy and identify saturated questions with rare failures. Second, for each candidate failure prefix, we estimate prefix-conditioned rollout accuracy to select the prefix closest to the target accuracy.

Table~\ref{tab:overhead} reports the token usage of each stage. The initial saturation-identification step uses 3.32B tokens when we sample 128 rollouts for every question. The prefix-selection step uses 1.94B tokens when we evaluate candidate prefixes on the full 10\%, 20\%, \(\ldots\), 90\% grid with 32 rollouts per prefix.

These costs can be reduced with simple approximations. For saturation identification, instead of sampling 128 rollouts for every question, we first sample 16 rollouts and only continue sampling for questions with 16/16 correct responses, stopping once an incorrect rollout is observed or the maximum of 128 attempts is reached. This staged procedure reduces the cost from 3.32B to 0.64B tokens, corresponding to 19.3\% of the original budget.

For prefix selection, we can use a coarser prefix grid and fewer rollouts per prefix. Specifically, replacing the 10\% grid with 20\% increments and reducing the number of rollouts from 32 to 8 lowers the cost from 1.94B to 0.11B tokens, or 5.8\% of the original budget. These results suggest that the computational overhead of failure-prefix conditioning can be substantially reduced without changing the overall training objective.

\section{Additional Experiment: Addressing Response Length Issue}
\label{sec:length}

\begin{table*}[h]
\centering
\caption{Performances on math benchmarks with controls for response length in RLVR training.}
\resizebox{\textwidth}{!}{
\begin{tabular}{lcccccc}
\toprule
\textbf{Model / Setting} & \textbf{MATH 500} & \textbf{AMC 12} & \textbf{AIME 24} & \textbf{AIME 25} & \textbf{HMMT 25} & \textbf{Average} \\
\midrule

Failure-Prefix, uncapped budget
& 86.6 & 60.5 & 33.1 & 27.1 & 15.4 & 44.5 \\

Failure-Prefix, prefix-counted budget
& 86.2 & 59.1 & 34.7 & 26.5 & 14.8 & 44.3 \\

\midrule

Medium data, 8k limit
& 85.7 & 58.8 & 34.2 & 26.2 & 15.0 & 44.0 \\

Medium data, 12k limit
& 86.2 & 59.5 & 35.1 & 25.8 & 15.2 & 44.4 \\

\bottomrule
\end{tabular}
}
\label{tab:length}
\end{table*}

All RLVR experiments in the paper use a response length limit of 8,000 tokens, as explained in Appendix~\ref{appendix:GRPO_training}. However, this raises two potential fairness concerns.

First, failure-prefix-conditioned examples receive a larger effective budget because generation begins after the prefix, while the prefix itself is not counted. To address this, we rerun failure-prefix conditioning with \(\tau=0.5\), counting the prefix length toward the 8,000-token limit and excluding the small number of prefixes that already exceed this limit. As shown in Table~\ref{tab:length}, the average accuracy is 44.3\%, close to the original 44.5\%. This suggests that the improvement comes from failure-prefix conditioning itself, rather than from a larger effective token budget.

Second, standard RLVR on medium-difficulty questions may be disadvantaged by the same 8,000-token limit, since these questions produce longer responses on average and are more likely to be truncated. To test this, we increase the response limit for the medium baseline to 12,000 tokens. Its average accuracy improves from 44.0\% to 44.4\%, showing that length matters for this baseline. However, this performance remains comparable to failure-prefix conditioning, so our conclusion is unchanged: applying failure-prefix conditioning to saturated problems is competitive with collecting new high-signal data.

\section{Full Results for Iterative Failure-Prefix Conditioning}
\label{appendix:iterative-results}

Table~\ref{tab:iterative_appendix} reports the benchmark-level results for the second iteration of failure-prefix conditioning described in Section~\ref{sec:round2}. Both variants improve over the first-iteration checkpoint, with the target-accuracy variant reaching the best average performance.

\begin{table}[h]
\centering
\caption{Benchmark-level performance after the second iteration of failure-prefix conditioning.}
\resizebox{\linewidth}{!}{
\begin{tabular}{lcccccc}
\toprule
\textbf{Model / Setting} & \textbf{MATH 500} & \textbf{AMC 12} & \textbf{AIME 24} & \textbf{AIME 25} & \textbf{HMMT 25} & \textbf{Average} \\
\midrule
Iteration 2 -- Target Accuracy ($\tau=0.5$) & 86.7 & 60.2 & 35.7 & 27.1 & 15.2 & 45.0 \\
Iteration 2 -- Fixed Truncation ($\gamma=0.5$) & 86.4 & 59.9 & 34.9 & 27.4 & 15.4 & 44.8 \\
\bottomrule
\end{tabular}
}
\label{tab:iterative_appendix}
\end{table}

\section{Derivation of Question-Level Weights and Their Relation to Reward Variance}
\label{app:weight-variance-derivation}

We provide a detailed derivation showing how question-level weights arise in GRPO-style objectives and why they are directly tied to the standard deviation of the reward distribution, following the analysis of~\citet{li2025disco}.

\paragraph{Setup}
We consider a binary reward setting where, for a given question $q$, each generated output $o$ receives a reward
\[
r(o \mid q) \in \{0,1\}.
\]
Let
\[
p(q) := \mathbb{E}_{o \sim \pi_{\text{old}}(\cdot \mid q)}[r(o \mid q)]
\]
denote the probability that the current policy $\pi_{\text{old}}$ produces a correct answer for question $q$.

Under this setting, the reward variance and standard deviation satisfy
\[
\mathrm{Var}[r(o \mid q)] = p(q)\bigl(1-p(q)\bigr),
\qquad
\mathrm{std}[r(o \mid q)] = \sqrt{p(q)\bigl(1-p(q)\bigr)}.
\]

\paragraph{Normalized advantage}
GRPO and its variants employ a normalized advantage function of the form
\[
A(o \mid q)
=
\frac{r(o \mid q) - p(q)}{\sqrt{p(q)(1-p(q))}}.
\]
Since the reward is binary, this admits the explicit piecewise representation
\[
A(o \mid q)
=
\begin{cases}
\sqrt{\dfrac{1-p(q)}{p(q)}}, & \text{if } r(o \mid q)=1,\\[8pt]
-\sqrt{\dfrac{p(q)}{1-p(q)}}, & \text{if } r(o \mid q)=0.
\end{cases}
\]

\paragraph{GRPO-style objective}
We consider the expectation form of the GRPO-family objective (omitting KL regularization for clarity):
\[
J_0(\theta)
=
\mathbb{E}_{q}
\mathbb{E}_{o \sim \pi_{\text{old}}(\cdot \mid q)}
\left[
\frac{1}{|o|}
\sum_{t=1}^{|o|}
f\!\left(
\frac{\pi_\theta(o_t \mid q, o_{<t})}
{\pi_{\text{old}}(o_t \mid q, o_{<t})},
\;
A(o \mid q)
\right)
\right].
\]

Here, $f(x,y)$ is a surrogate objective function inherited from PPO-style policy optimization.
For GRPO specifically,
\[
f(x,y) = \min\bigl(xy, \operatorname{clip}(x,1-\epsilon,1+\epsilon)\,y\bigr),
\]
where $x$ is a likelihood ratio and $y$ is the advantage.

\paragraph{Structural assumption on $f$}
Following~\citet{li2025disco}, we assume that $f(x,y)$ is non-decreasing in $x$ and admits the decomposition
\[
f(x,y)
=
\mathbf{1}\{y>0\}\, y\, f^{+}(x,1)
\;-\;
\mathbf{1}\{y\le 0\}\, |y|\, f^{-}(x,1),
\]
for some non-decreasing functions $f^{+}$ and $f^{-}$.
This assumption holds for GRPO and its common variants, and separates the effect of the sign and magnitude of the advantage.

\paragraph{Decomposition by reward outcomes.}
Let $\pi^+_{\text{old}}(\cdot \mid q)$ and $\pi^-_{\text{old}}(\cdot \mid q)$ denote the conditional distributions of outputs given $r(o \mid q)=1$ and $r(o \mid q)=0$, respectively.
Applying the law of total expectation and substituting the piecewise form of $A(o \mid q)$ yields
\begin{align*}
J_0(\theta)
&=
\mathbb{E}_{q}
\Biggl[
p(q)\sqrt{\frac{1-p(q)}{p(q)}}
\mathbb{E}_{o \sim \pi^+_{\text{old}}(\cdot \mid q)}
\frac{1}{|o|}
\sum_{t=1}^{|o|}
f^{+}\!\left(
\frac{\pi_\theta}{\pi_{\text{old}}},1
\right) \\
&\qquad\quad
-
(1-p(q))\sqrt{\frac{p(q)}{1-p(q)}}
\mathbb{E}_{o \sim \pi^-_{\text{old}}(\cdot \mid q)}
\frac{1}{|o|}
\sum_{t=1}^{|o|}
f^{-}\!\left(
\frac{\pi_\theta}{\pi_{\text{old}}},1
\right)
\Biggr].
\end{align*}

\paragraph{Emergence of the variance-based weight}
Factoring out common terms, the objective simplifies to
\[
J_0(\theta)
=
\mathbb{E}_{q}
\sqrt{p(q)\bigl(1-p(q)\bigr)}
\left[
\mathbb{E}_{o \sim \pi^+_{\text{old}}(\cdot \mid q)} s_\theta(o,q)
-
\mathbb{E}_{o \sim \pi^-_{\text{old}}(\cdot \mid q)} s_\theta(o,q)
\right],
\]
where $s_\theta(o,q)$ denotes the corresponding token-averaged scoring function.

Therefore, each question $q$ contributes to the objective with a multiplicative weight
\[
\omega(q) = \sqrt{p(q)\bigl(1-p(q)\bigr)} = \mathrm{std}[r(o \mid q)].
\]

\paragraph{Implication}
This derivation shows that GRPO-style objectives implicitly scale each question by the standard deviation of its reward distribution. As a result, questions that are nearly always correct ($p(q)\approx 1$) or nearly always incorrect ($p(q)\approx 0$) receive vanishing weight, while questions of intermediate difficulty are emphasized.

%%%%%%%%%%%%%%%%%%%%%%%%%%%%%%%%%%%%%%%%%%%%%%%%%%%%%%%%%%%%

\newpage

\end{document}